\documentclass[10pt,twocolumn,letterpaper]{article}

\usepackage[pagenumbers]{cvpr} 

\newcommand{\red}[1]{{\color{red}#1}}

\usepackage{booktabs}   
\usepackage{multirow}   
\usepackage{graphicx}   
\usepackage{amssymb}    
\usepackage[dvipsnames,table]{xcolor}
\usepackage{arydshln} 

\newcommand{\gridimg}[1]{%
  \raisebox{-0.5\height}{\includegraphics[width=0.21\textwidth]{#1}}%
}
\usepackage{tabularx}
\usepackage{pifont} 
\newcommand{\cmark}{{\color{ForestGreen}\ding{52}}}
\newcommand{\xmark}{{\color{red}\ding{55}}}
\definecolor{slategray1}{HTML}{e6f2ff} 

\newcommand{\rowlabel}[2]{%
  $\vcenter{\hbox{\rotatebox{90}{%
    \begin{tabular}{c}
      \textbf{#1} \\
      \footnotesize (#2)
    \end{tabular}%
  }}}$%
}
\newcommand{\rotatedtitle}[1]{$\vcenter{\hbox{\rotatebox{90}{\textbf{#1}}}}$}
\definecolor{darkgrayline}{gray}{0.35}

\usepackage{tikz}

\newcommand{\insertimg}[1]{\raisebox{-0.5\height}{\includegraphics[width=0.2\textwidth]{#1}}}
\usepackage{stfloats}
\usepackage{tablefootnote}

\definecolor{cvprblue}{rgb}{0.21,0.49,0.74}
\usepackage[pagebackref,breaklinks,colorlinks,allcolors=cvprblue]{hyperref}
\iftoggle{cvprfinal}{}{\hypersetup{pdfauthor={}, pdftitle={}}}
\usepackage[capitalize]{cleveref}

\def\paperID{488} 
\def\confName{3DV\xspace}
\def\confYear{2027\xspace}

\title{RawSLAM: Online HDR Gaussian SLAM from Linear Radiance}

\author{Marina Orozco González \qquad Luis Merino\\
{\tt\small \{morogon, lmercab\}@upo.es}
}

\begin{document}
\twocolumn[{%
  \renewcommand\twocolumn[1][]{#1}%
  \maketitle
  \vspace{-1.5em}
  \begingroup
    \captionsetup{type=figure}
    \centering
\newlength{\rightcolw}
\setlength{\rightcolw}{0.44\textwidth}
\newlength{\labelw}
\setlength{\labelw}{1.2em}
\newlength{\colgap}
\setlength{\colgap}{2.5pt}
\newlength{\rowgap}
\setlength{\rowgap}{3.5pt}
\newlength{\headgap}
\setlength{\headgap}{1.5pt}
\newlength{\imgw}
\setlength{\imgw}{0.31\dimexpr\rightcolw - \labelw\relax}

\newsavebox{\sampleimg}
\savebox{\sampleimg}{%
    \setlength{\fboxsep}{0pt}%
    \setlength{\fboxrule}{0.5pt}%
    \fbox{\includegraphics[width=\dimexpr\imgw-2\fboxrule\relax, keepaspectratio]{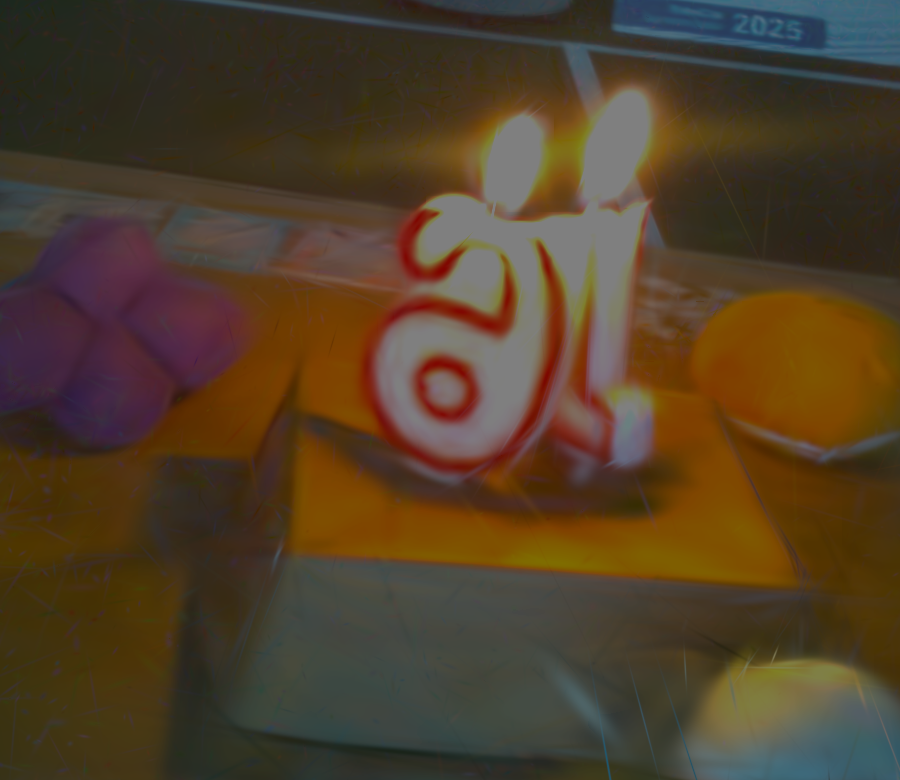}}%
}
\newlength{\singleimgh}
\setlength{\singleimgh}{\dimexpr\ht\sampleimg+\dp\sampleimg\relax}

\newsavebox{\rightcontent}
\savebox{\rightcontent}{%
    \setlength{\fboxsep}{0pt}%
    \setlength{\fboxrule}{0.5pt}%
    \hbox{%
        \begin{minipage}[b]{\labelw}%
            \centering
            \makebox[\linewidth]{\small\vphantom{$\mathbf{\times 0.5}$}}%
            \par\vspace{\headgap}%
            \vbox to \singleimgh{\vss\hbox to \linewidth{\hss\rotatedtitle{LDR}\hss}\vss}%
            \par\vspace{\rowgap}%
            \vbox to \singleimgh{\vss\hbox to \linewidth{\hss\rotatedtitle{HDR}\hss}\vss}%
        \end{minipage}%
        \hspace{\colgap}%
        \begin{minipage}[b]{\imgw}%
            \centering
            \small $\mathbf{\times 0.5}$\par\vspace{\headgap}%
            \usebox{\sampleimg}\par\vspace{\rowgap}%
            \fbox{\includegraphics[width=\dimexpr\linewidth-2\fboxrule\relax, keepaspectratio]{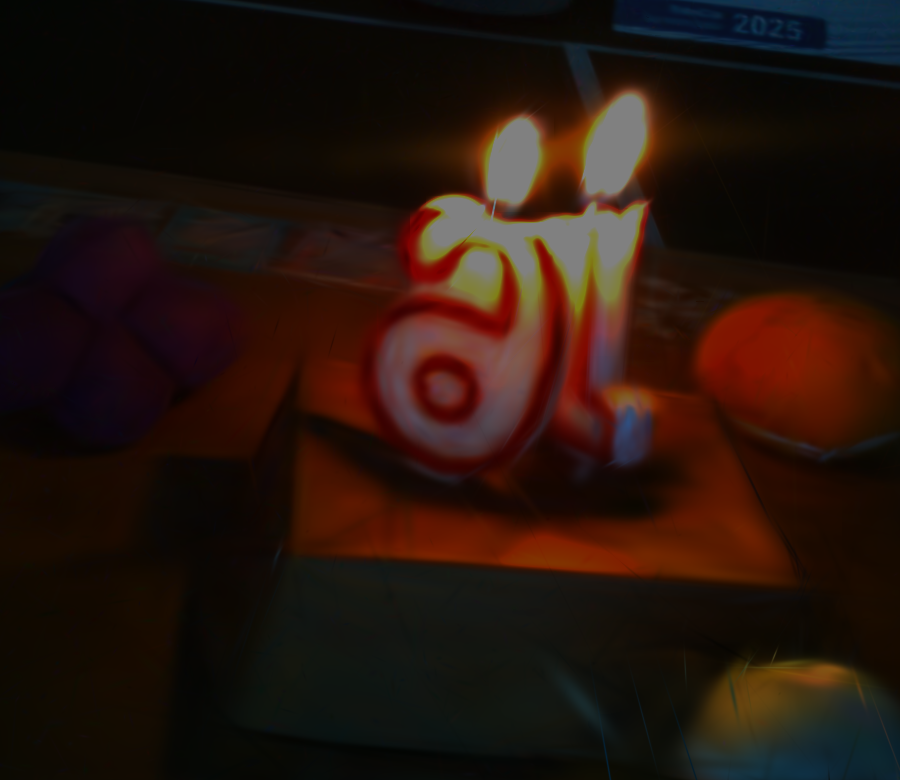}}%
        \end{minipage}%
        \hspace{\colgap}%
        \begin{minipage}[b]{\imgw}%
            \centering
            \small $\mathbf{\times 1.0}$\par\vspace{\headgap}%
            \fbox{\includegraphics[width=\dimexpr\linewidth-2\fboxrule\relax, keepaspectratio]{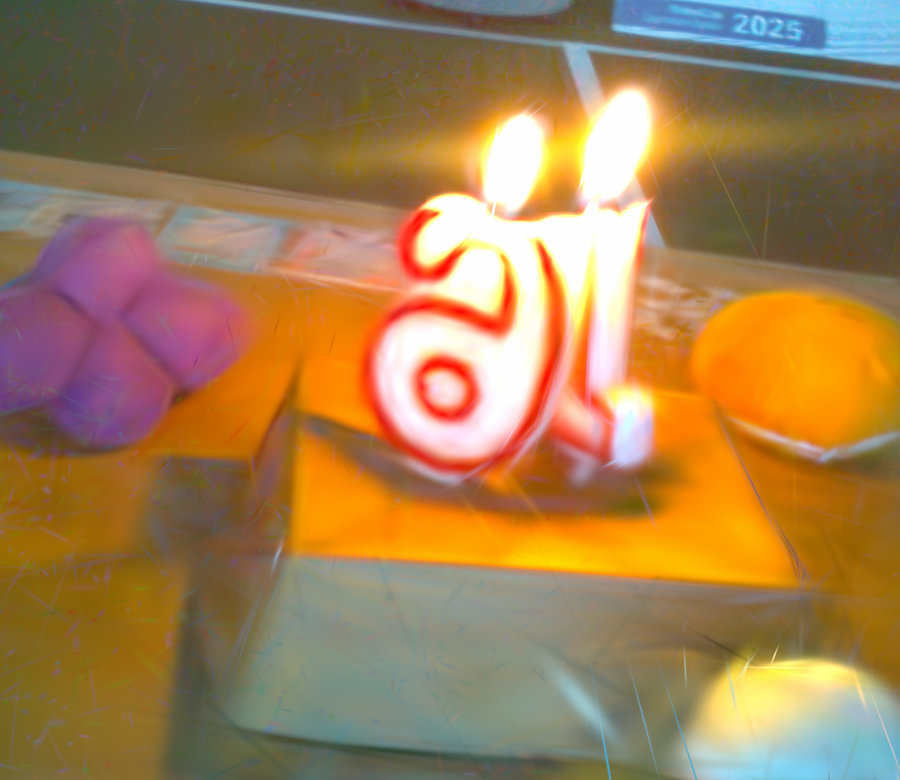}}\par\vspace{\rowgap}%
            \fbox{\includegraphics[width=\dimexpr\linewidth-2\fboxrule\relax, keepaspectratio]{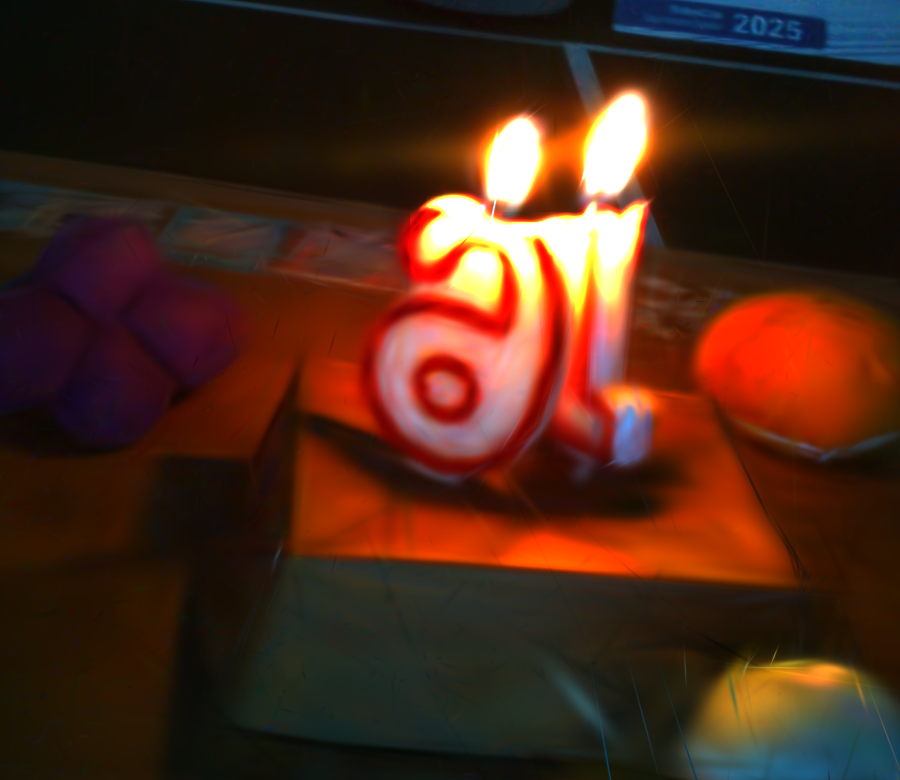}}%
        \end{minipage}%
        \hspace{\colgap}%
        \begin{minipage}[b]{\imgw}%
            \centering
            \small $\mathbf{\times 4.0}$\par\vspace{\headgap}%
            \fbox{\includegraphics[width=\dimexpr\linewidth-2\fboxrule\relax, keepaspectratio]{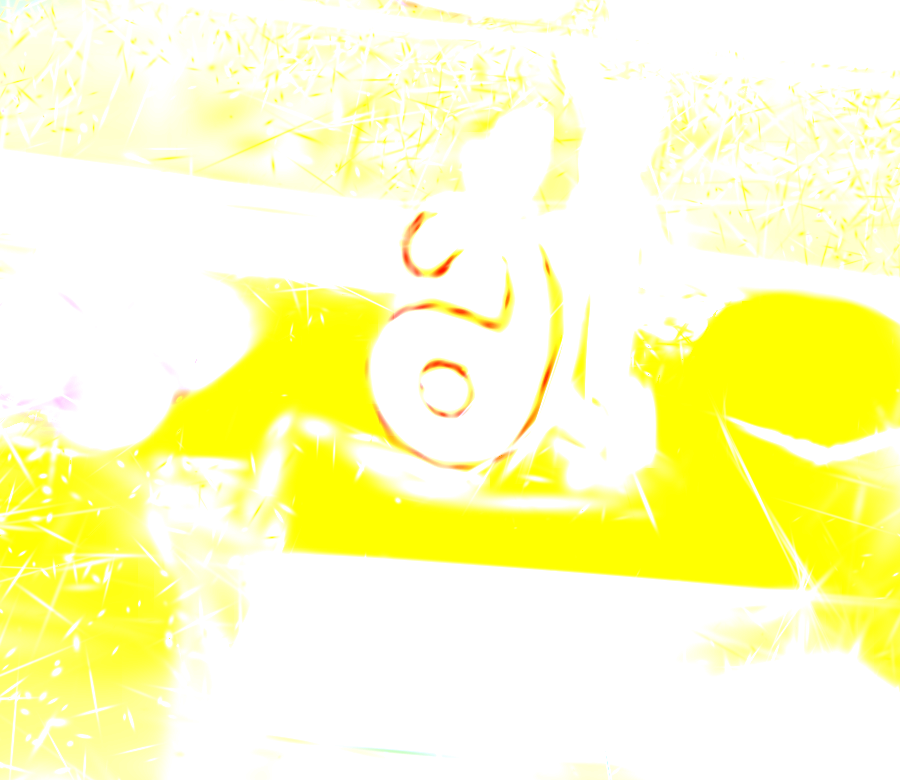}}\par\vspace{\rowgap}%
            \fbox{\includegraphics[width=\dimexpr\linewidth-2\fboxrule\relax, keepaspectratio]{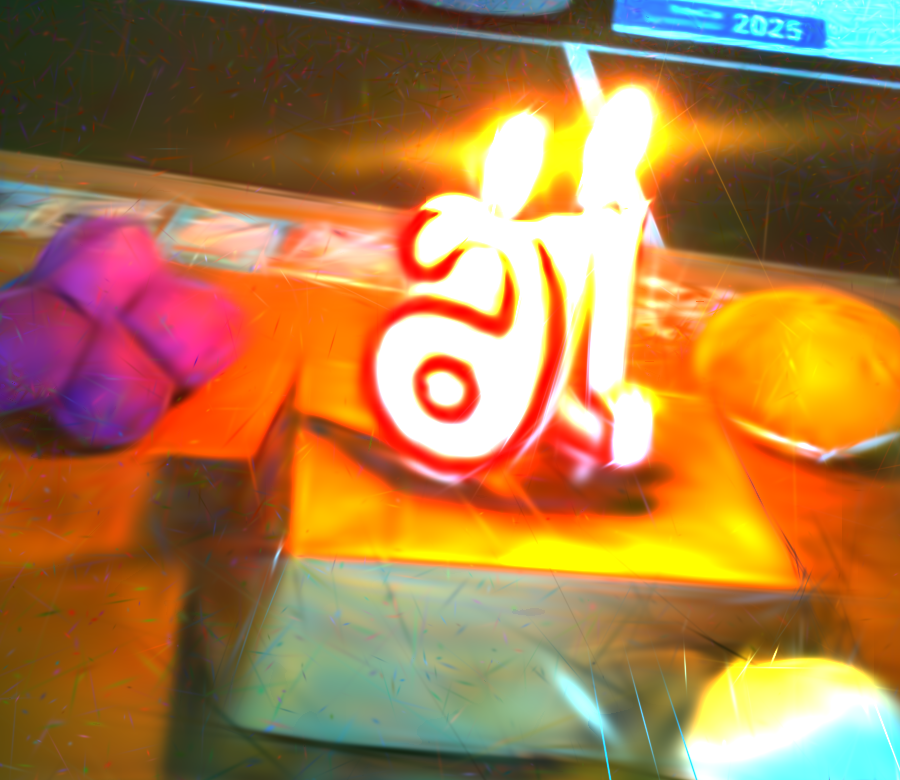}}%
        \end{minipage}%
    }%
}

\newsavebox{\leftimg}
\savebox{\leftimg}{%
    \includegraphics[height=\dimexpr\ht\rightcontent+\dp\rightcontent\relax, keepaspectratio]{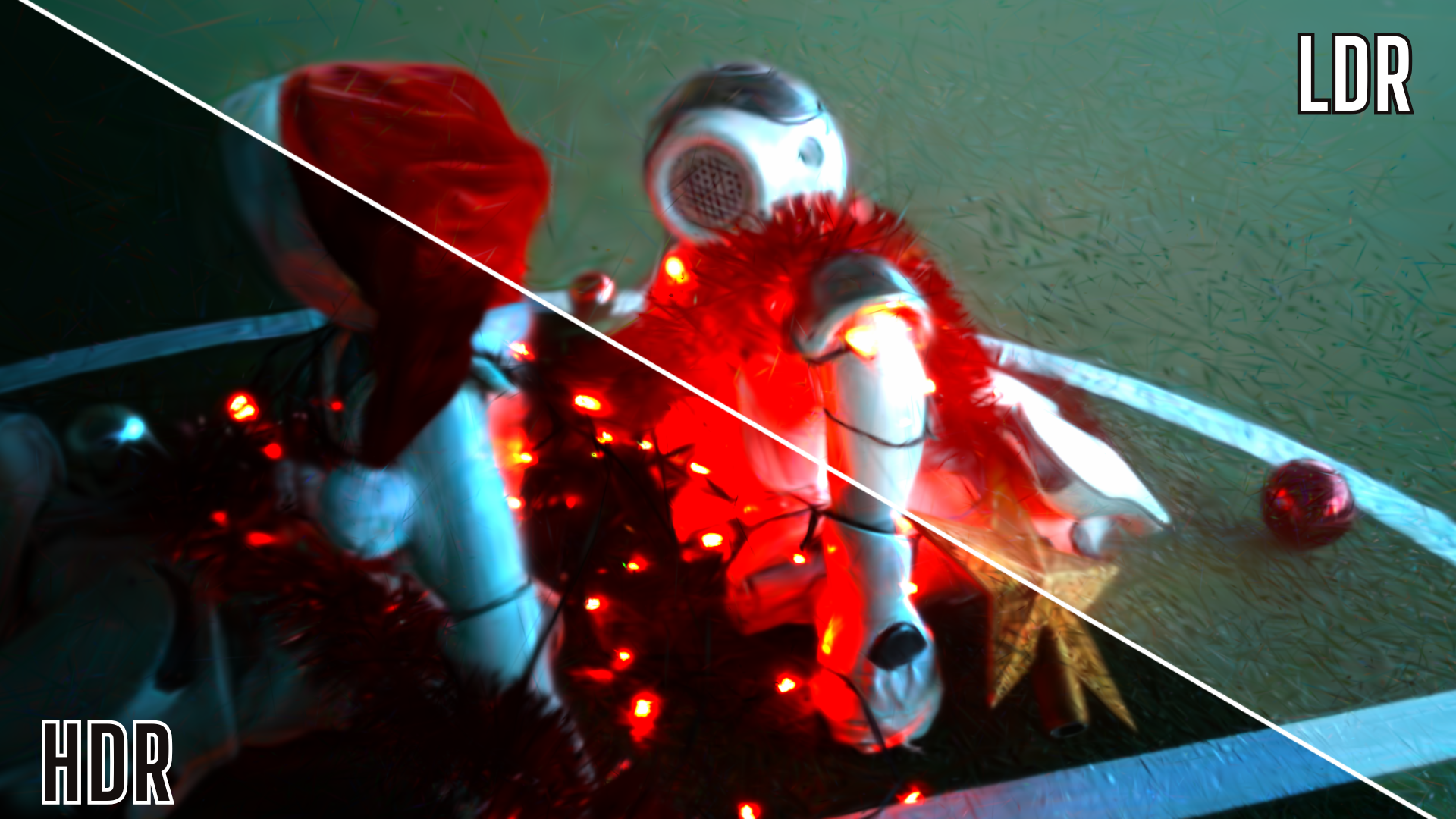}%
}

\resizebox{\textwidth}{!}{%
    \begin{subfigure}[b]{\wd\leftimg}%
        \raggedleft
        \raisebox{-\dp\rightcontent}{\usebox{\leftimg}}%
        \caption{HDR and LDR renders}
        \label{subfig:main}
    \end{subfigure}%
    \hspace{1.0em}%
    \begin{subfigure}[b]{1.0pt}%
        \centering
        \raisebox{-\dp\rightcontent}{%
            \begin{tikzpicture}
                \draw[dashed, line width=0.6pt, gray!90] (0,0) -- (0,\dimexpr\ht\rightcontent+\dp\rightcontent\relax);
            \end{tikzpicture}%
        }%
        \caption*{\phantom{(a)}}%
    \end{subfigure}%
    \hspace{1.0em}%
    \begin{subfigure}[b]{\wd\rightcontent}%
        \raggedright
        \usebox{\rightcontent}%
        \caption{Exposure Variation}
        \label{subfig:exposures}
    \end{subfigure}%
}

\caption{
Our pipeline jointly estimates camera poses and reconstructs high-fidelity 3DGS representations directly from 16-bit linear HDR image sequences. (a) Reconstructed linear HDR rendering alongside its tonemapped sRGB LDR counterpart. (b) Virtual exposure sweep applied directly to our rendered HDR views, demonstrating post-capture exposure adjustment without highlight clipping and shadow quantization inherent to conventional LDR pipelines.  Views are rendered at native full resolution to showcase peak reconstruction fidelity.
}

\label{fig:main}
\vspace{1.5em}
  \endgroup
  \vspace{1.5em}
}]
\begin{abstract}

Current dense visual SLAM systems rely almost exclusively on 8-bit tonemapped Low Dynamic Range (LDR) inputs, limiting their robustness in extreme lighting where shadows and highlights trigger tracking drift and mapping collapse. Conversely, existing raw and High Dynamic Range (HDR) reconstruction pipelines operate strictly offline. They depend on Structure-from-Motion preprocessing and are not suited for large inter-frame motion. We present, to the best of our knowledge, the first online Gaussian SLAM framework that tracks and maps directly on single-exposure 16-bit linear HDR imagery. Our method rests on three core components: an architecture-agnostic HDR Gaussian Splatting module featuring an MLP-free logarithmic parameterization of Gaussian color features; a Reinhard range-compressed photometric objective; and structure-guided spatial gradient weighting. Combined, these components allow our approach to outperform a direct HDR adaptation of MonoGS in both trajectory and reconstruction accuracy, while rendering natively in linear scene radiance for post-rendering processing. The same formulation runs unchanged on standard 8-bit inputs, roughly halving the MonoGS baseline error. Furthermore, our HDR Gaussian module transfers seamlessly to SplaTAM, Gaussian SLAM, and DROID-W, eliminating all tracking failures these systems suffer on challenging illumination sequences. To enable this research, we introduce RawSLAM: a dataset of 10 real-world indoor sequences featuring 16-bit RAW imagery, aligned depth, IMU measurements, and external OptiTrack poses. Code and dataset will be made publicly available soon.

\end{abstract}    
\section{Introduction}
\label{sec:intro}

\begin{figure*}[t!]
    \centering
    \includegraphics[width=0.8\linewidth]{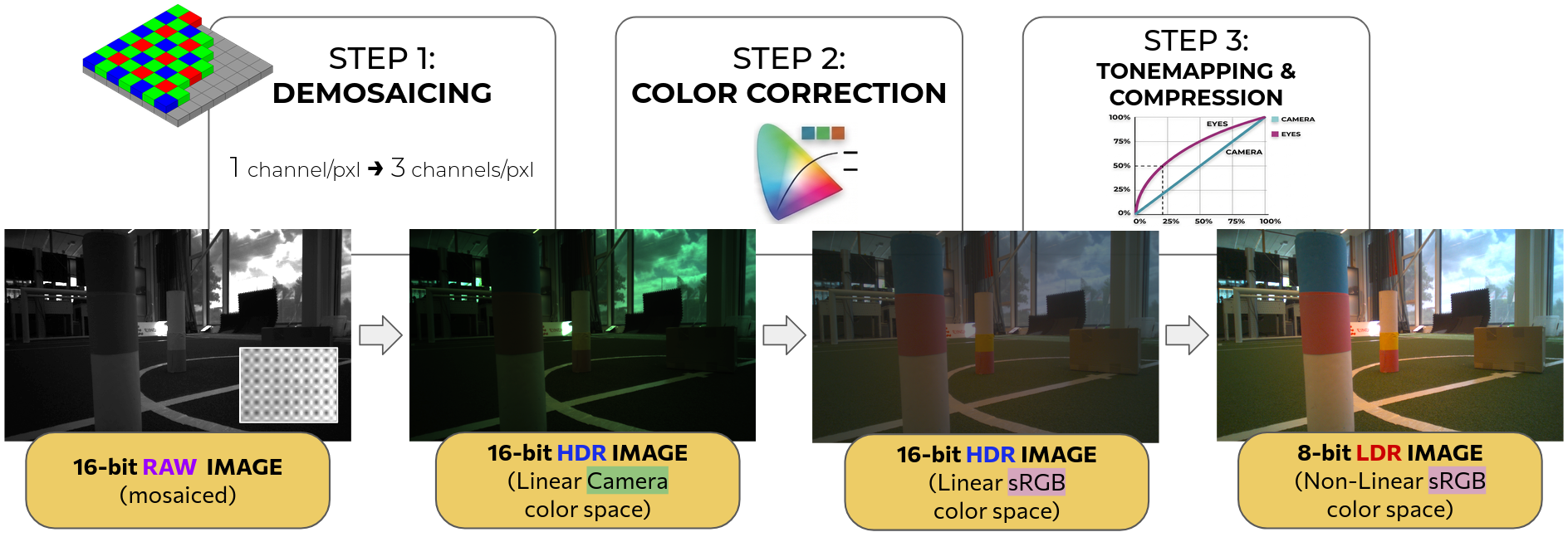}
    \caption{
    Simplified overview of a standard camera ISP pipeline, illustrating the primary processing stages of RAW sensor data.
    (1) \textit{\textbf{Demosaicing}} converts the single-channel RAW image into a three-channel, full-size RGB image by interpolating the missing values in the Bayer pattern based on neighboring pixels in the color filter array.
    (2) To produce a visually balanced image, a \textit{\textbf{color correction matrix}} is typically used to convert the image from its camera-specific color space to a standardized one, such as sRGB. (3) Through \textit{\textbf{tonemapping}}, the linear relation that pixel values have with the perceived light is lost, applying a gamma compression that assigns more bits to darker values, while the use of 8 bits as final resolution compresses the wider depth that raw sensors contain. In this work, our experiments are run over 16-bit linear sRGB HDR images and with 8-bit non-linear sRGB LDR images.}
    \label{fig:image_processing_pipeline}
\end{figure*}

Low Dynamic Range (LDR) images are the most common type of images that digital cameras provide. However, the information that the raw sensors in the camera capture from the received light is typically transformed through an internal Image Signal Processing (ISP) pipeline \cite{seeinthedark, unprocessing} that irreversibly clips extreme highlights and introduces non-linear transformations that lose luminance information (\Cref{fig:image_processing_pipeline}). In contrast, linear High Dynamic Range (HDR) images retain scene radiance information over a substantially wider dynamic range than conventional LDR images. By bypassing lossy tone mapping, non-linear gamma compression, and 8-bit quantization \cite{reinhard, mantiuk}, linear HDR sensor data retains crucial geometric and photometric details in both deep shadows and extreme highlights.

Recent advances in offline 3D reconstruction have successfully explored the use of HDR data to tackle extreme lighting and noise. Works like RawNeRF \cite{rawnerf} pioneered the use of unbounded linear RAW data within a NeRF \cite{NeRF} framework, while subsequent methods like HDR-NeRF \cite{hdrnerf}, HDR-GS \cite{hdrgs}, HDRSplat \cite{HDRSplat} and LE3D \cite{le3d} adapted similar concepts to accelerate training and rendering. 
However, a key limitation of these approaches is that they operate offline and rely on pre-computed camera poses from multi-step Structure-from-Motion (SfM) pipelines like COLMAP \cite{colmap}, and are brittle to large wide-baseline camera movements.

Conversely, dense Simultaneous Localization and Mapping (SLAM) systems (e.g., MonoGS \cite{monogs}, SplaTAM \cite{splatam}, Gaussian SLAM \cite{gaussianslam}, DROID-W \cite{droidw}) achieve simultaneous online tracking and 3D reconstruction, but they predominantly rely on LDR images.

Our main goal in this work is to bridge this gap and evaluate the behavior of Gaussian SLAM systems when processing HDR images directly from standard camera sensors.

To enable this, we introduce a new dataset: RawSLAM. Comprising 10 scenes averaging 1,860 frames per scene recorded in challenging lighting scenarios, RawSLAM provides RAW image files (from which both LDR and HDR streams are extracted), synchronized depth and IMU information, and highly accurate ground-truth camera poses recorded via an external OptiTrack motion capture system. We also open-source the complete pipeline required to process these RAW files into usable HDR and LDR streams.

Using this dataset, we demonstrate that current Gaussian SLAM methods, when applied as is, cannot cope with high dynamic lighting scenes, resulting in failed tracking. Furthermore, even when modified to accept HDR inputs, these vanilla architectures are unable to fully leverage the potential of HDR data. While a few recent works like \citet{sdgs} have attempted to bring HDR benefits into SLAM, they rely on spatial differences generated by specialized high-frame-rate hybrid-pixel hardware rather than processing full HDR colors.

To address these limitations, we introduce a highly adaptable, architecture-agnostic HDR module that plugs into existing 3DGS SLAM frameworks with minimal changes to enable online HDR rendering. Crucially, to ensure optimization stability across the dynamic range of linear scene radiance, this pluggable module incorporates a direct, MLP-free logarithmic parameterization of the Gaussian color features natively. 

To exploit this representation during training, we introduce an HDR-aware loss that combines Reinhard dynamic-range compression with structure-guided spatial gradient weights. Crucially, these contributions are color-space agnostic: this stable log-parameterization and edge-aware loss not only prevent optimization collapse under native HDR inputs, but also generalize exceptionally well to standard LDR streams, consistently enhancing the tracking and mapping performance of conventional sRGB SLAM baselines.

Our main contributions can be summarized as follows:

\begin{enumerate}
    \item \textbf{Novel Dataset}: We introduce RawSLAM, a comprehensive SLAM dataset featuring RAW images, depth, IMU data, and externally computed ground-truth camera poses, alongside an open-source pipeline for RAW-to-HDR/LDR extraction.
    \item \textbf{Architecture-Agnostic Framework}: We present the first implementation of online HDR training and rendering for Gaussian SLAM that can be integrated into existing frameworks and operates on data from standard camera sensors capable of recording in RAW or HDR formats.
    \item \textbf{Optimized Parameterization and Loss}: We implement a feature parameterization and loss formulation specifically designed for linear HDR rendering, substantially improving tracking and rendering performance for both HDR and LDR inputs.
\end{enumerate}


\section{Related Work}
\label{sec:related_work}

\subsection{Offline and Online Novel-View synthesis} 
Achieving novel-view synthesis from multi-view images requires representing the scene with elements that capture fine details across perspectives while remaining memory-efficient.
The introduction of NeRF \cite{NeRF} shifted the paradigm of novel-view synthesis toward coordinate-based neural representations optimized by gradient descent. Its variants introduced strategies like depth supervision \cite{depthsupervision1, depthsupervision2, depthsupervision3}, or distortion loss \cite{distortionlossnerf}. However, the photorealism achieved by NeRF relies on dense volumetric ray marching through a multi-layer perceptron (MLP), which bottlenecks rendering speed. To address this limitation, 3D Gaussian Splatting (3DGS) \cite{gaussiansplatting} replaces ray marching with explicit scene representation, using unstructured 3D Gaussians whose parameters (position, scale, opacity, covariance, color) are optimized during gradient descent.

However, these are inherently offline methods that require known camera poses. Traditionally, these poses are either obtained from static external devices in a controlled environment, which severely limits the range of reconstructable scenarios, or computed via a heavy pre-processing step using Structure-from-Motion (SfM) pipelines \cite{sfm1, sfm3}, with \citet{colmap} being the most widely used. To overcome this, SLAM methods enable sequential reconstruction while jointly estimating the camera trajectory. 

Visual SLAM systems historically rely on diverse representations, including sparse features \cite{orbslam2, orbslam3}, direct photometric tracking \cite{dsoslam, lsdslam}, dense grids \cite{kinectfusion}, surfels \cite{elasticfusion}, and neural implicit frameworks \cite{imap, coslam, voxfusion, eslam}.
Recently, the integration of 3D Gaussian Splatting into SLAM systems, such as SplaTAM \cite{splatam}, Gaussian SLAM \cite{gaussianslam}, and most notably MonoGS \cite{monogs}, among others \cite{gaus-slam, pointslam, fgsslam, wildgs, photoslam, cgslam, mgslam}, has achieved state-of-the-art results in dense reconstruction. Concurrently, emerging literature is focusing on SLAM models over dynamic scenarios, as typically SLAM methods assume a static scene. 
To handle real-world noise and distractors, some methods incorporate deep visual features \cite{droidslam, tartanvo, sgsslam}, such as DROID-W \cite{droidw}.

However, these methods only accept LDR sRGB data for training. Consequently, these systems cannot recover an HDR scene representation suitable for downstream HDR rendering operations such as exposure adjustment, white balancing and tone mapping. In contrast, our module enables Gaussian SLAM to train and render HDR images.

\subsection{HDR images in 3D Reconstruction} 
HDR images are of significant interest in computational photography, as they preserve the true radiometric information captured by the camera sensor. In contrast, standard LDR images have undergone non-linear tone mapping and dynamic range compression, irreversibly discarding the sensor's original data. Because HDR images bypass these lossy transformations in the digital imaging pipeline, they retain a significantly broader dynamic range, making them invaluable for professional post-processing \cite{mantiuk, banterle, saraswat, reinhard, ldr2hdr3}. The fundamental importance of these uncompressed data is further evidenced by the extensive body of literature dedicated to inverting the camera pipeline to recover HDR data from LDR inputs \cite{ldr2hdr1, ldr2hdr2, ldr2hdr3, ldr2hdr4}.

To preserve the fidelity of captured light data, researchers in computational photography have turned to modeling scenes in an unbounded linear color space. RawNeRF \cite{rawnerf} pioneered this approach by optimizing a neural radiance field directly on noisy, linear RAW input images, demonstrating that operating in RAW space preserves the full dynamic range and naturally averages out zero-mean sensor noise. Subsequent methods, such as HDR-NeRF \cite{hdrnerf} and HDR-GS \cite{hdrgs}, extended 3D reconstruction to high-dynamic-range scenes. However, instead of taking true RAW or HDR files as input, these approaches rely on sets of multi-exposure LDR images to estimate an implicit Camera Response Function (CRF) and reconstruct the underlying HDR radiance field. 
To achieve direct rendering from linear RAW inputs, HDRSplat \cite{HDRSplat} and LE3D \cite{le3d} extend this paradigm to 3D Gaussian Splatting, with the latter replacing standard Spherical Harmonics (SH) with a lightweight, exponentially activated color MLP.

Crucially, all these methods are strictly offline and depend on pre-computed camera poses from SfM algorithms like COLMAP \cite{colmap}. The closest existing approach to incorporating HDR information into online SLAM is SDGS \cite{sdgs}, which feeds spatial differences (gradients) derived from HDR data into a 3DGS tracking pipeline. However, SDGS relies on spatial differences generated by specialized hybrid-pixel hardware (a Tianmouc camera \cite{tianmouc}), making it difficult to generalize to standard camera setups. In contrast, our method directly processes linear HDR images obtained from any standard camera capable of capturing RAW or HDR files, enabling online HDR novel-view synthesis and dense mapping in a live SLAM pipeline without requiring pre-computed poses.

\section{Method}
\label{sec:method}


Our goal is to enable Gaussian SLAM in high-contrast environments by leveraging the full radiance information preserved in linear 16-bit HDR images. Standard SLAM pipelines rely on tonemapped 8-bit sRGB inputs, which discard critical luminance data in shadows and highlights. To adapt our baseline SLAM framework, MonoGS \cite{monogs}, to linear, high-dynamic-range color spaces, we introduce an HDR-aware Gaussian SLAM framework and a specialized photometric optimization scheme.


\subsection{HDR Gaussian Splatting module}

\begin{figure}[t!]
    \centering
    \includegraphics[width=\linewidth]{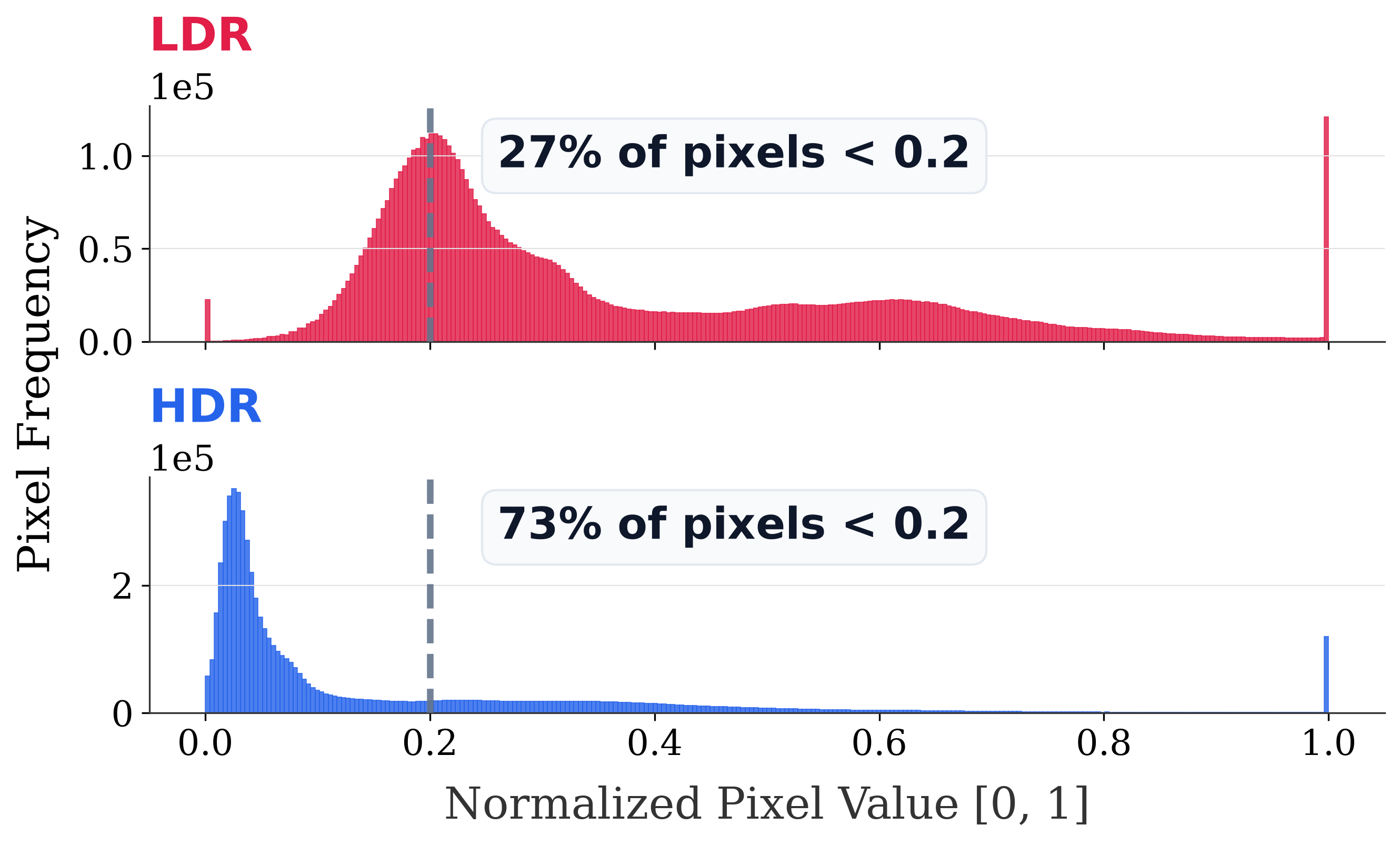}
    \caption{Pixel values distribution in tonemapped LDR images (top) and in linear HDR color space images (bottom).}
    \label{fig:pxl_value_distribution}
\end{figure}

Standard Gaussian Splatting representations, whether utilizing Spherical Harmonics (SH) or direct view-independent RGB features, are fundamentally optimized for non-linear tonemapped sRGB color spaces.  Consequently, even when normalized, they struggle with linear RAW and HDR pixel intensities due to the highly skewed distribution they exhibit, where shadow and midtone information is compressed into small numerical values (\Cref{fig:pxl_value_distribution}). 

To leverage the full physical dynamic range of HDR radiance in SLAM without the computational and memory overhead of auxiliary color MLPs, used by offline methods such as LE3D \cite{le3d}, we implement a direct logarithmic parameterization of the Gaussian color features.

Let $c_i$ denote the linear color of Gaussian primitive $i$, parameterized by its color feature $f_i$. 
When primitive $i$ is instantiated from pixel $p$ in keyframe $k$, let $c_{p}^{k}$ denote the color associated with the pixel from which it is back-projected. We initialize the feature as $f_i = \log(\max(c_{p}^{k}, \epsilon))$, with $\epsilon=10^{-6}$, and replace the bounded rendering activations with an exponential mapping, $c_i = \exp(f_i)$, during rasterization. Applying the chain rule, the optimization gradient of the training loss $\mathcal{L}$ with respect to the log-space feature $f_i$ of primitive $i$ is expressed as:
\begin{equation}
    \frac{\partial \mathcal{L}}{\partial f_i} = \frac{\partial \mathcal{L}}{\partial c_i} \cdot \exp(f_i) = \frac{\partial \mathcal{L}}{\partial c_i} \cdot c_i
    \label{eq:exp_param}
\end{equation}



Since Adam \cite{adam} normalizes each update by the running magnitude of that parameter's gradient, the step in $f_i$ has roughly constant size regardless of $c_i$, and in log space such a step is a percentage change in color.
Dark primitives ($c_i \to 0$) thus receive small absolute updates that cannot overshoot into the midtones, while their relative step matches that of a bright primitive, allocating optimization precision uniformly across the orders of magnitude spanned by linear HDR data and counteracting the skewed value distribution of \Cref{fig:pxl_value_distribution}.
Positivity is also enforced by construction: $\exp(f_i) > 0$, so primitives never rely on the zero-clamp of standard 3DGS, which yields dead gradients for primitives driven negative during tracking and mapping.

While \citet{hdrgs} and \citet{le3d} implement exponential color activations via an MLP, in our supplementary experiments we show that a direct logarithmic parameterization is substantially more efficient for Gaussian SLAM.




\subsection{Loss function}
Standard photometric losses in Gaussian SLAM typically apply an $L_1$ penalty directly on LDR sRGB pixel intensities. 
However, a plain $L_1$ loss on linear HDR values is poorly conditioned once composed with \Cref{eq:exp_param}: since $\partial \mathcal{L} / \partial f_i$ scales with $c_i$, a bright primitive receives gradients orders of magnitude larger than a shadow primitive for the same absolute residual. 

Inspired by the differentiable range-compression strategy introduced by \citet{rawnerf}, we propose a tone-mapped photometric objective that embeds a differentiable Reinhard operator \cite{reinhard} within our $L_1$ loss:
\begin{equation}
    \mathcal{L}_{1-Reinhard} = \sum_p \Bigl| T(I_{R,p})-T(I_{GT,p}) \Bigr|
\end{equation}
\noindent where $I_{R,p}$ and $I_{GT,p}$ denote the linear radiance of pixel $p$ in the rendered and ground-truth images, respectively, and $T(x) = x/{(1+x)}$ is the Reinhard tonemapping operator. Differentiating this objective with respect to a Gaussian color feature $f_i$ yields:
\begin{equation}
    \frac{\partial \mathcal{L}}{\partial f_i} = \sum_p  \frac{\mathrm{sign} \bigl( T(I_{R,p}) - T(I_{GT,p}) \bigr)}{(1+I_{R,p})^2} \cdot \frac{\partial I_{R,p}}{\partial f_i}
\end{equation}

Note that $ \partial I_{R,p} \,/ \,\partial f_i$ directly incorporates the proportional scaling in \Cref{eq:exp_param}.  
The gradient shrinkage factor $1/(1+I_{R,p})^2$ naturally dampens the influence of extremely bright pixels as rendered values increase. Conversely, mid-tones receive balanced updates, and dark shadow pixels receive strong, uncompressed gradients toward the ground-truth. This ensures more uniform sensitivity across the entire luminance spectrum compared to plain $L_1$ on linear HDR.

\paragraph{HDR-Aware Spatial Gradient Weighting}
In high-contrast environments, textureless or flat regions can yield ambiguous photometric signals that degrade camera tracking stability. Inspired by the gradient supervision strategy of \citet{sdgs}, we apply a normalized spatial gradient weight $w(p)$ to modulate the photometric loss across the image plane:


\begin{equation}
w(p) = \frac{|\nabla I_{GT}(p)|}{\Bigl( \max_q \bigl( |\nabla I_{GT}(q)| \bigr) + \varepsilon \Bigr)} + \delta_{\text{floor}}
\end{equation}

\noindent where $|\nabla I_{GT}(p)| = \sqrt{g_v^2(p) + g_h^2(p)}$ represents the gradient magnitude at pixel $p$ computed via vertical ($g_v$) and horizontal ($g_h$) Scharr filter convolutions, $q$ indexes all pixel locations to determine the global image maximum, and $\varepsilon$ is a constant that we set to $1e^{-6}$.

We set $\delta_{\text{floor}} = 0.1$, constraining $w(p) \in [0.1,\ 1.1]$. 
Consequently, high-gradient edge pixels are weighted up to $1.1$, while uniform regions are down-weighted to $0.1$. Maintaining a non-zero floor prevents flat regions from being fully masked, ensuring the tracking objective does not degenerate into a sparse, edge-only alignment.


In standard LDR pipelines, gamma compression and quantization often crush low-light details into uniform black regions, whereas linear color spaces preserve subtle gradient structures and shadow boundaries. By computing $w(p)$ directly on the linear HDR intensity, our weighting scheme exploits the full dynamic range of the signal to identify structurally informative edges that would otherwise be lost in conventional LDR representations.

Because spatial gradient weighting is designed to prevent pose drift by prioritizing structurally informative edges, we apply the modulation weight $w(p)$ exclusively during the camera tracking phase. Applying $w(p)$ during mapping would down-weight uniform surfaces, causing Gaussian primitives on smooth walls or flat areas to under-converge. 

\paragraph{Tracking and Mapping Photometric Objectives}
To further enforce structural consistency across extreme exposure variations, we incorporate a Structural Similarity (SSIM) dissimilarity term. While this term is a standard component of the photometric loss in our LDR baseline, standard SSIM is perceptually calibrated for bounded low dynamic range color spaces and is incompatible with linear HDR radiance. 
Consequently, prior to evaluating SSIM, we apply the same tonemapping pipeline used to generate baseline LDR images from RAW files to both the rendered and ground-truth HDR images. Our final combined photometric objectives for camera tracking $\mathcal{L}^{\text{pho}}_{\text{T}}$ and Gaussian mapping $\mathcal{L}^{\text{pho}}_{\text{M}}$ are thus defined as:


\begin{equation}
    \mathcal{L}^{\text{pho}}_{\text{T}} =  \lambda \cdot \mathcal{L}_{\text{SSIM}} +(1-\lambda) \cdot \sum_p  w(p) \, \mathcal{L}_{1-Reinhard}^p 
    \label{eq:totalloss}
\end{equation}

\begin{equation}
    \mathcal{L}^{\text{pho}}_{\text{M}} =   \lambda \cdot \mathcal{L}_{\text{SSIM}} +(1-\lambda) \cdot \mathcal{L}_{1-Reinhard} 
    \label{eq:mapping_loss}
\end{equation}

\noindent where $\mathcal{L}_{\text{SSIM}} = \frac{1}{2} \Bigl( 1 - \text{SSIM} \bigl(I^{\text{LDR}}_{R}, I^{\text{LDR}}_{GT}\bigr) \Bigr) $, and $\lambda=0.2$ balances the contribution of spatial structural similarity against pixel-wise luminance accuracy.

\subsection{Implementation details}

Our system is built upon the MonoGS \cite{monogs} codebase, retaining its core camera tracking, keyframe management, and windowed sub-map optimization architecture. To support 16-bit linear HDR streams, we introduce targeted modifications across the data pipeline, CUDA rasterization kernel, and evaluation protocol.

\paragraph{Data Pipeline \& Ingestion.}
Standard Gaussian Splatting libraries and data loaders are hardcoded for 8-bit LDR images normalized by $255$. To preserve the dynamic range of linear RAW and HDR inputs without precision loss, we extend the data ingestion pipeline to natively load \texttt{uint16} images normalized by $65535$. 

\paragraph{Modified CUDA Rasterization Kernel.}

The standard 3D Gaussian Splatting CUDA rasterizer assumes 8-bit outputs and applies an alpha culling threshold, $\alpha_{\text{threshold}}$, of $1/255$, below which a contribution cannot alter the rendered pixel.
As rendering to 16-bit images lowers the smallest representable increment to $1/65535$, we modify the underlying CUDA rasterization kernel to lower $\alpha_{\text{threshold}}$ accordingly for 16-bit inputs.
This preserves the contribution, and the gradient, of low-opacity Gaussians whose effect is invisible at 8 bits but well above the 16-bit floor, typically those carrying faint structure in dark scene regions.

\paragraph{Generalizability to LDR sRGB Inputs.}
While our logarithmic color parameterization, Reinhard-compressed photometric loss, and spatial gradient weighting are motivated by the dynamic range of linear HDR radiance, our formulation processes either LDR or HDR intensities. Since all core algorithmic modifications operate natively on normalized intensities, our implementation seamlessly switches between 8-bit and 16-bit processing depending on the input. Consequently, the exact same pipeline applies directly to standard LDR sequences without architectural changes. 
Interestingly, our results show that this unified approach notably enhances both tracking and rendering.

\section{Experiments}
\label{sec:experiments}
\subsection{Experimental Setup}
\paragraph{Dataset.}
Due to the scarcity of visual SLAM benchmarks providing long-sequence linear RAW imagery with ground-truth trajectories, we evaluate our method on \textbf{RawSLAM}, a novel dataset of 10 indoor sequences (averaging $\sim$1,860 frames per scene, $>\!18{,}600$ frames total) recorded using an Intel RealSense D435i sensor and saved as 16-bit raw DNG files. Alongside visual streams, the dataset provides aligned depth maps and IMU measurements. The sequences capture diverse and extreme indoor lighting conditions, including direct sunlight, high-contrast shadows, transparent objects, and bright light sources, to rigorously test tracking stability. Synchronized 6-DoF ground-truth camera poses were acquired via a millimeter-accurate OptiTrack motion capture system. To enable fair comparisons against standard baselines, RawSLAM provides both unprocessed 16-bit linear HDR streams and ISP-processed 8-bit sRGB LDR streams.

To assess its generalization capability across standard LDR domains, we also evaluate our LDR pipeline on the TUM RGB-D benchmark \cite{tum}.

\paragraph{System Setup.}

We evaluate our system on a desktop with an Intel Core i7-14700K CPU and an NVIDIA Quadro RTX 5000 GPU, downsampling input frames to $360 \times 640$. To ensure full reproducibility and eliminate dependency ambiguity across different host environments, the entire pipeline, evaluation scripts, and modified CUDA kernels are containerized using Docker and released publicly along with the RawSLAM dataset. We maintain identical hyperparameter configurations across all comparable experiments.

\paragraph{Metrics.}
We evaluate tracking accuracy using the Root Mean Square Error (RMSE) of the Absolute Trajectory Error (ATE), keeping alignment procedures consistent with the tested baselines.
To evaluate map quality, we follow \citet{rawnerf} to decouple optimization from evaluation. For perceptual benchmarking, rendered linear images undergo percentile-based exposure rescaling and sRGB gamma compression before computing standard LDR metrics ($\text{PSNR}$, $\text{SSIM}$, $\text{LPIPS}$). However, because tonemapping masks physical radiometric errors, we adopt $\text{PSNR-}\mu$ ($\mu = 5000$) \cite{pu-psnr} as our primary HDR fidelity benchmark. By applying logarithmic $\mu$-law compression directly to the linear radiance, $\text{PSNR-}\mu$ prevents extreme specular highlights from disproportionately dominating the photometric error over dark shadow regions. 

We report the average across three runs for
all our evaluations. In our tables, we highlight up to the \colorbox[HTML]{99E4B0}{\textbf{first}}, \colorbox[HTML]{d8fcdb}{\underline{second}}, and \colorbox[HTML]{fffbda}{third} best results for HDR runs, and the \colorbox[HTML]{FF9999}{\textbf{first}}, \colorbox[HTML]{FCDFD8}{\underline{second}}, and \colorbox[HTML]{fffbda}{third} best for LDR runs.


\subsection{Quantitative Evaluation}

\begin{table}[t]
\centering
\caption{\textbf{Quantitative results across scenes}. Besides PSNR-$\mu$, all rendering metrics are computed over the tonemapped version of the HDR renders. Vanilla HDR stands for the HDR adaptation of MonoGS without any of our further changes. Numbers in red {(\red{N})} indicate the number of scene tracking failures, the mean is computed among the successful runs. }
\label{tab:results}
\renewcommand{\arraystretch}{1.2}
\footnotesize
\resizebox{\linewidth}{!}{%
\begin{tabular}{lcccccc}
\toprule
& \begin{tabular}{@{}c@{}} \textbf{ATE RMSE} \\ (cm) $\downarrow$\end{tabular} 
 & \begin{tabular}{@{}c@{}}\textbf{PSNR-$\mu$} \\ (dB) $\uparrow$\end{tabular} 
 & \begin{tabular}{@{}c@{}}\textbf{PSNR} \\ (dB) $\uparrow$\end{tabular} 
 & \begin{tabular}{@{}c@{}}\textbf{SSIM} \\ $\uparrow$\end{tabular} 
 & \begin{tabular}{@{}c@{}}\textbf{LPIPS} \\ $\downarrow$\end{tabular} 
 & \begin{tabular}{@{}c@{}}\textbf{Depth $L_1$} \\ (cm) $\downarrow$\end{tabular} \\ \midrule
SplaTAM \cite{splatam} (\red{\textbf{3}}) &  89.29 & - & \cellcolor[HTML]{fffbda}19.13 & \cellcolor[HTML]{FCDFD8}\underline{0.706} & \cellcolor[HTML]{FCDFD8}\underline{0.405} & \cellcolor[HTML]{FCDFD8}\underline{104.83} \\
Gaussian SLAM \cite{gaussianslam} (\red{\textbf{1}}) & \cellcolor[HTML]{fffbda} 86.93 & - & \cellcolor[HTML]{FF9999}\textbf{21.86} & \cellcolor[HTML]{FF9999}\textbf{0.803} & \cellcolor[HTML]{FF9999}\textbf{0.384} & \cellcolor[HTML]{FF9999}\textbf{55.46} \\
MonoGS \cite{monogs} & \cellcolor[HTML]{FCDFD8}\underline{56.84} & - & 18.36 & 0.483 & 0.569 & 154.72 \\
Ours (LDR) & \cellcolor[HTML]{FF9999}\textbf{25.97} & - & \cellcolor[HTML]{FCDFD8}\underline{20.12} & \cellcolor[HTML]{fffbda}0.548 & \cellcolor[HTML]{fffbda}0.433 & \cellcolor[HTML]{fffbda}107.32 \\
\cmidrule(l){1-7}
Vanilla HDR & \cellcolor[HTML]{d8fcdb}\underline{49.43} & \cellcolor[HTML]{d8fcdb}\underline{20.96} & \cellcolor[HTML]{d8fcdb}\underline{17.33} & \cellcolor[HTML]{99E4B0}\textbf{0.529} & \cellcolor[HTML]{d8fcdb}\underline{0.570} & \cellcolor[HTML]{d8fcdb}\underline{159.36} \\
Ours (HDR) & \cellcolor[HTML]{99E4B0}\textbf{24.15} & \cellcolor[HTML]{99E4B0}\textbf{22.96} & \cellcolor[HTML]{99E4B0}\textbf{18.92} & \cellcolor[HTML]{d8fcdb}\underline{0.515} & \cellcolor[HTML]{99E4B0}\textbf{0.450} & \cellcolor[HTML]{99E4B0}\textbf{118.10} \\
\bottomrule
\end{tabular}%
}
\end{table}

\paragraph{Camera Tracking Accuracy}

As reported in Table~\ref{tab:results}, when operating natively on linear HDR inputs, \mbox{Ours (HDR)} achieves the lowest overall ATE of $24.15\text{ cm}$. This represents a $51.1\%$ reduction in trajectory error compared to the direct baseline adaptation (\mbox{Vanilla HDR} at $49.43\text{ cm}$), demonstrating that our explicit log-space color parameterization and photometric gradient weighting are essential to effectively exploit HDR information within the tracking optimization. Furthermore, when operating on standard LDR inputs, \mbox{Ours (LDR)} reduces the tracking error of the baseline MonoGS architecture by nearly half ($25.97\text{ cm}$ vs. $56.84\text{ cm}$) and significantly outperforms heavier dense visual SLAM frameworks such as SplaTAM ($89.29\text{ cm}$) and Gaussian SLAM ($86.93\text{ cm}$).

While challenging scenes with moving shadows, specularities, and low exposure cause catastrophic tracking divergence in other systems, with SplaTAM and Gaussian SLAM failing to track $\red{3}$ and $\red{1}$ sequences entirely, respectively, our method successfully completes all evaluation trajectories without a single failure, exhibiting exceptional tracking robustness under extreme lighting variations.

Moreover, comparing our own configurations reveals that natively processing linear HDR radiance yields consistent tracking improvements over the LDR pipeline across almost all individual sequences (we provide a detailed per-scene ATE breakdown in the Supplementary Material).

\begin{table}[b]
\centering
\caption{\textbf{ATE RMSE} $\downarrow$ (cm) on the\textbf{ LDR TUM \cite{tum} dataset.}}
\label{tab:tum_results}
\footnotesize 
\resizebox{0.8
 \linewidth}{!}{%
    \begin{tabular}{lccc|c}
    \toprule
    Config & fr1/desk  & fr2/xyz & fr3/office& \textbf{Mean} \\ \midrule
    Baseline & 2.09 & 1.29 & 2.13 & 1.84 \\
    Ours & \cellcolor[HTML]{99E4B0} \textbf{1.80} & \cellcolor[HTML]{99E4B0} \textbf{1.19} & \cellcolor[HTML]{99E4B0} \textbf{1.92} & \cellcolor[HTML]{99E4B0} \textbf{1.64} \\
    \bottomrule 
    \end{tabular} 
}
\end{table}

\paragraph{Generalization to standard LDR benchmarks.} 

To verify generalization to conventional LDR imagery, in \Cref{tab:tum_results} we evaluate three TUM RGB-D \cite{tum} sequences (\texttt{fr1/desk}, \texttt{fr2/xyz}, \texttt{fr3/office}). Our method reduces the baseline mean ATE RMSE from $1.84\text{ cm}$ to $1.64\text{ cm}$, indicating that our modifications do not compromise performance when trained with standard LDR inputs on conventional, well-lit sequences.

\begin{figure*}[]
  \centering
  \small
  \setlength{\tabcolsep}{2pt}
  
  \begin{tabular}{c c c c c}
    & \textbf{Baseline (LDR)} & \textbf{Ours (LDR)} & \textbf{Ours (HDR)} & \textbf{Ground-truth} \\[6pt]
    
    \rowlabel{coat\_rack}{$1911$} &
    \insertimg{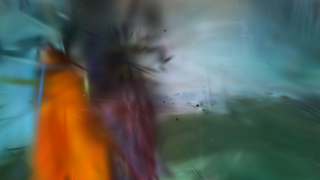} &
    \insertimg{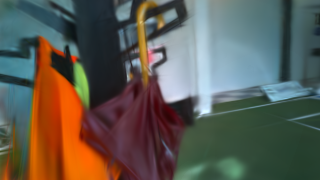} &
    \insertimg{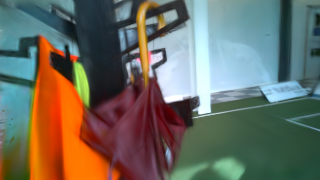} &
    \insertimg{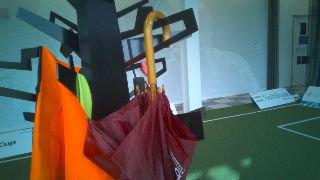} \\[28.5pt]  
    
    \rowlabel{bottles}{1937} &
    \insertimg{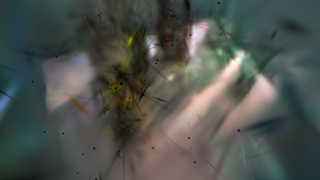} &
    \insertimg{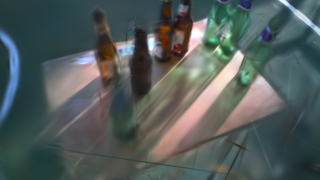} &
    \insertimg{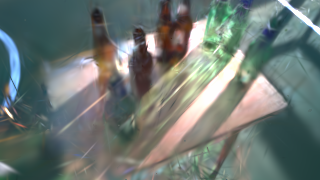} &
    \insertimg{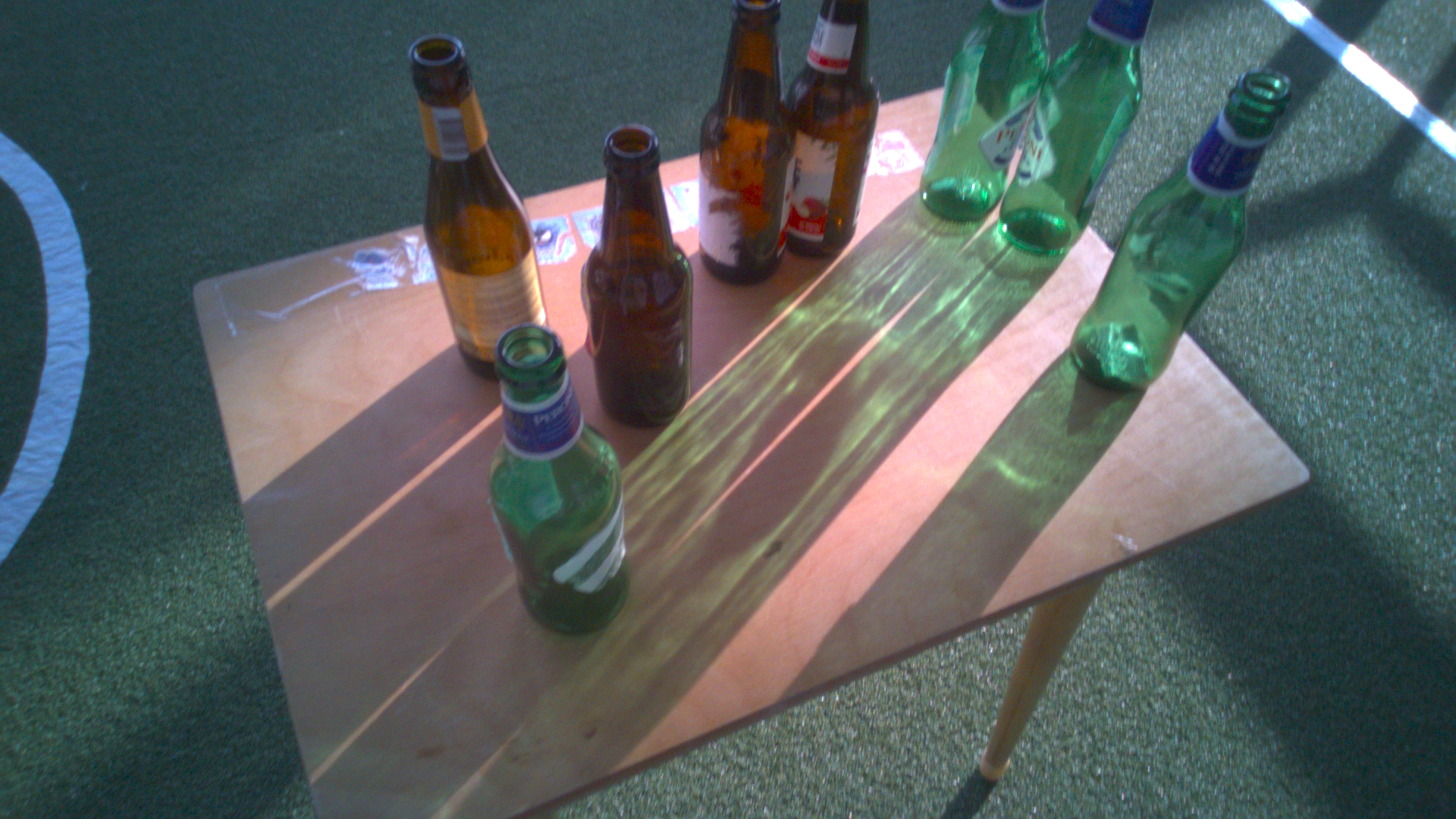} \\[28.5pt]  
    
    \rowlabel{coffee}{1916} &
    \insertimg{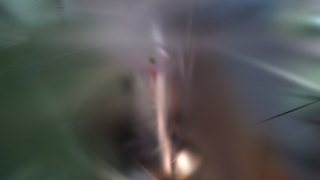} &
    \insertimg{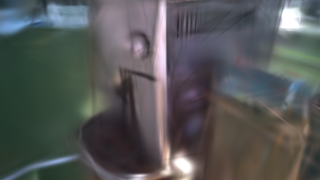} &
    \insertimg{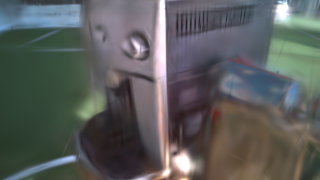} &
    \insertimg{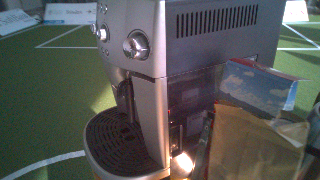} \\  
  \end{tabular}

  \caption{\textbf{Rendering performance} for three long sequences of the RawSLAM dataset. Scene lengths are in parentheses. Results in Ours (HDR) have been converted to LDR sRGB color space for fair comparison.}
  \label{fig:three_renders}
\end{figure*}
\paragraph{Rendering Fidelity} 
As shown in Table~\ref{tab:results}, \mbox{Ours (HDR)} achieves a $+2.00\text{ dB}$ gain in $\text{PSNR-}\mu$ over \mbox{Vanilla HDR} ($22.96\text{ dB}$ vs. $20.96\text{ dB}$), demonstrating that our log-space formulation accurately reconstructs the HDR radiance field. On tonemapped sRGB evaluations, our method consistently improves PSNR in both LDR ($+1.76\text{ dB}$ over MonoGS) and HDR ($+1.59\text{ dB}$ over Vanilla HDR) modes, while reducing geometric mapping error ($\text{Depth } L_1$) by over $40\text{ cm}$ in both configurations. Although SplaTAM and Gaussian SLAM achieve higher rendering metrics, they exhibit substantially larger trajectory errors on our challenging sequences (tracking divergence and $86\text{--}89\text{ cm}$ ATE on their successful runs). In contrast, our pipeline ensures joint trajectory stability and accurate physical reconstruction.

\paragraph{Computational Efficiency and Runtime}
We measure end-to-end system throughput under concurrent tracking and mapping on a single GPU. Under standard LDR configurations, our pipeline sustains $1.29\text{ FPS}$, representing a $16.2\%$ speedup over the MonoGS baseline ($1.11\text{ FPS}$). Similarly, in HDR mode, \mbox{Ours (HDR)} achieves $1.22\text{ FPS}$, outperforming \mbox{Vanilla HDR} ($1.08\text{ FPS}$) by $13.0\%$. These results confirm that our formulation consistently improves system throughput. By maintaining stable tracking, we avoid the computational overhead of optimizing redundant Gaussians in drifted views.

\subsection{Qualitative Evaluation}
We present qualitative reconstruction results for three representative scenarios in \Cref{fig:three_renders}: high-contrast areas with bright windows, reflective metallic surfaces, and transparent glass bottles. While the LDR baseline fails to reconstruct these challenging scenes, yielding highly blurry visualizations, our pipeline succeeds in both LDR and HDR configurations. Crucially, several intricate details are reconstructed with even higher fidelity thanks to our linear HDR representation. More examples of renders of the remaining seven sequences of the dataset are provided in the supplementary material.


Because hardware limitations require us to downsample the inputs to $360 \times 640$ for our long and highly dynamic trajectories, we additionally visualize our reconstruction on short, stable segments at the $1080 \times 1920$ native resolution. \Cref{fig:main} shows the resulting reconstruction quality and demonstrates that our formulation can preserve fine details and highlights when operating at the native input resolution.

\subsection{Plug-and-Play HDR GS module}
In \Cref{tab:across_models}, we demonstrate the plug-and-play nature of our approach by integrating our HDR GS module into three dense photometric Gaussian Splatting architectures: MonoGS \cite{monogs}, SplaTAM \cite{splatam}, and Gaussian SLAM \cite{gaussianslam}. We also tested our module on DROID-W \cite{droidw}, a state-of-the-art deep bundle-adjustment pipeline. 

Our module integrates seamlessly into these diverse backbones, reducing mean ATE by 32--39\% across the three Gaussian Splatting systems while leaving DROID-W essentially unchanged ($16.30 \rightarrow 16.46$~cm).
These gains are lower bounds: each LDR mean covers only the sequences that system completes, whereas each HDR mean covers all 10 scenes.
Our formulation eliminates every tracking failure ($\red{3}$ for SplaTAM, $\red{1}$ for Gaussian SLAM and DROID-W) where baseline LDR models collapsed in all runs of challenging illumination scenes: \textit{candles} for SplaTAM and Gaussian SLAM, \textit{coffee} for SplaTAM and DROID-W, and \textit{small\_city} for SplaTAM.

Surprisingly, DROID-W, whose deep feature extractor is trained on LDR images, tracked the HDR \textit{coffee} sequence while failing on its LDR version, suggesting the potential benefits of HDR in learning-based tracking models.



Finally, we also tested the Vanilla HDR version of our module on a traditional sparse feature-based system, ORB-SLAM2 \cite{orbslam2}, where we obtained a negligible tracking difference of $+0.40\text{ cm}$ with respect to the $4.00\text{ cm}$ ATE from the baseline, 
confirming the broad applicability of our HDR module across diverse SLAM architectures.


\begin{table}
\centering
\caption{\textbf{Tracking }(ATE RMSE)\textbf{ and rendering }(PSNR, SSIM, LPIPS, PSNR-$\mu$)\textbf{ metrics across SLAM frameworks, comparing the LDR baselines against our version of them with our HDR GS module}. We report the mean across scenes and runs. Numbers in red {(\red{N})} indicate the number of scene tracking failures, the mean is computed among the successful runs. 
}
\label{tab:across_models}
\renewcommand{\arraystretch}{1.2}
\resizebox{\linewidth}{!}{%
\begin{tabular}{llccccc}
\toprule
\textbf{Model} & & \begin{tabular}[c]{@{}c@{}}\textbf{ATE RMSE}\\(cm) $\downarrow$\end{tabular} & \begin{tabular}[c]{@{}c@{}}\textbf{PSNR}\\(dB) $\uparrow$\end{tabular} & \begin{tabular}[c]{@{}c@{}} \textbf{SSIM}\\ $\uparrow$  \end{tabular} & \begin{tabular}[c]{@{}c@{}}\textbf{LPIPS}\\ $\downarrow$ \end{tabular}& \begin{tabular}[c]{@{}c@{}}\textbf{PSNR}-$\mu$\\(dB) $\uparrow$\end{tabular} \\
\midrule
\multirow{2}{*}{DROID-W \cite{droidw}} & LDR {(\red{\textbf{1}})} & 16.30 & 18.56 & 0.533 & 0.529 & - \\
 & \cellcolor{slategray1}HDR & \cellcolor{slategray1} 16.46 & \cellcolor{slategray1} 15.99 & \cellcolor{slategray1} 0.421 & \cellcolor{slategray1} 0.612 & \cellcolor{slategray1} 19.43 \\
 \midrule
\multirow{2}{*}{Gaussian SLAM \cite{gaussianslam}} & LDR {(\red{\textbf{1}})} & 86.93 & 21.86 & 0.803 & 0.384 & - \\
 & \cellcolor{slategray1}HDR & \cellcolor{slategray1} 57.28 & \cellcolor{slategray1} 19.78 & \cellcolor{slategray1} 0.720 & \cellcolor{slategray1} 0.529 & \cellcolor{slategray1} 20.71 \\
\cdashline{1-7}
\multirow{2}{*}{SplaTAM \cite{splatam}} & LDR {(\red{\textbf{3}})}  & 89.29 & 19.13 & 0.706 & 0.405 & - \\
 & \cellcolor{slategray1}HDR & \cellcolor{slategray1} 54.26 & \cellcolor{slategray1} 16.74 & \cellcolor{slategray1} 0.581 & \cellcolor{slategray1} 0.519 & \cellcolor{slategray1} 19.12 \\
\cdashline{1-7}
\multirow{2}{*}{MonoGS \cite{monogs}} & LDR & 56.84 & 18.36 & 0.483 & 0.569 & - \\
 & \cellcolor{slategray1}HDR & \cellcolor{slategray1}38.47 & \cellcolor{slategray1}18.30 & \cellcolor{slategray1}0.480 & \cellcolor{slategray1}0.494 & \cellcolor{slategray1}23.12 \\

\bottomrule
\end{tabular}%
}
\end{table}

\begin{table*}[b!]
\centering
\caption{\textbf{Ablation study} of different configuration components on LDR and HDR outputs across 3 runs and across all the scenes. The asterisk * marks that the baseline version of LDR includes a $L_{\text{SSIM}}$ term, while the vanilla HDR version does not include it.}
\label{tab:ablation}
\renewcommand{\arraystretch}{1.2}
\resizebox{\linewidth}{!}{%
\begin{tabular}{ccccccccccccc}
\toprule
\multicolumn{4}{c}{\textbf{Config}} & \multicolumn{4}{c}{\textbf{LDR}} & \multicolumn{5}{c}{\textbf{HDR}} \\
\cmidrule(lr){1-4} \cmidrule(lr){5-8} \cmidrule(lr){9-13}
+ Weight & + $L_{\text{SSIM}}$ & + L1-Reinhard & + log reparam. & ATE RMSE (cm)$\downarrow$ & PSNR (dB)$\uparrow$ & SSIM$\uparrow$ & LPIPS$\downarrow$ & ATE RMSE (cm)$\downarrow$ & PSNR (dB)$\uparrow$ & SSIM$\uparrow$ & LPIPS$\downarrow$ & PSNR-$\mu$ (dB)$\uparrow$ \\ \midrule
\makebox[1.2em][c]{\xmark} & \makebox[1.2em][c]{*} & \makebox[1.2em][c]{\xmark} & \makebox[1.2em][c]{\xmark} & 56.84 & 18.36 & 0.483 & 0.569 & 49.43 & 17.33 & \cellcolor[HTML]{99E4B0}\textbf{0.529} & 0.570 & 20.96 \\
\makebox[1.2em][c]{\xmark} & \makebox[1.2em][c]{*} & \makebox[1.2em][c]{\xmark} & \makebox[1.2em][c]{\textbf{\cmark}} & 41.19 & 19.60 & 0.524 & \cellcolor[HTML]{fffbda}0.459 & 38.47 & 18.30 & 0.480 & 0.494 & \cellcolor[HTML]{d8fcdb}\underline{23.12} \\
\makebox[1.2em][c]{\xmark} & \makebox[1.2em][c]{*} & \makebox[1.2em][c]{\textbf{\cmark}} & \makebox[1.2em][c]{\xmark} & 30.11 & \cellcolor[HTML]{fffbda}19.84 & \cellcolor[HTML]{FCDFD8}\underline{0.542} & \cellcolor[HTML]{FCDFD8}\underline{0.458} & 28.49 & 18.34 & 0.469 & 0.526 & 22.41 \\
\makebox[1.2em][c]{\xmark} & \makebox[1.2em][c]{*} & \makebox[1.2em][c]{\textbf{\cmark}} & \makebox[1.2em][c]{\textbf{\cmark}} & 30.60 & 19.65 & \cellcolor[HTML]{fffbda}0.536 & 0.460 & \cellcolor[HTML]{d8fcdb}\underline{26.74} & \cellcolor[HTML]{d8fcdb}\underline{18.91} & \cellcolor[HTML]{fffbda}0.487 & \cellcolor[HTML]{fffbda}0.482 & \cellcolor[HTML]{fffbda}23.06 \\ \cdashline{1-13}
\makebox[1.2em][c]{\xmark} & \makebox[1.2em][c]{\xmark} & \makebox[1.2em][c]{\textbf{\cmark}} & \makebox[1.2em][c]{\textbf{\cmark}} & \cellcolor[HTML]{fffbda}28.76 & 19.76 & 0.514 & 0.483 & \cellcolor[HTML]{d8fcdb}\underline{26.74} & \cellcolor[HTML]{d8fcdb}\underline{18.91} & \cellcolor[HTML]{fffbda}0.487 & \cellcolor[HTML]{fffbda}0.482 & \cellcolor[HTML]{fffbda}23.06 \\
\makebox[1.2em][c]{\textbf{\cmark}} & \makebox[1.2em][c]{\xmark} & \makebox[1.2em][c]{\textbf{\cmark}} & \makebox[1.2em][c]{\textbf{\cmark}} & \cellcolor[HTML]{FCDFD8}\underline{28.67} & \cellcolor[HTML]{FCDFD8}\underline{19.89} & 0.516 & 0.472 & \cellcolor[HTML]{fffbda}27.67 & \cellcolor[HTML]{fffbda}18.76 & 0.483 & 0.489 & 22.83 \\
\makebox[1.2em][c]{\xmark} & \makebox[1.2em][c]{\textbf{\cmark}} & \makebox[1.2em][c]{\textbf{\cmark}} & \makebox[1.2em][c]{\textbf{\cmark}} & 30.60 & 19.65 & \cellcolor[HTML]{fffbda}0.536 & 0.460 & 29.88 & 18.66 & \cellcolor[HTML]{d8fcdb}\underline{0.515} & \cellcolor[HTML]{d8fcdb}\underline{0.452} & \cellcolor[HTML]{99E4B0}\textbf{23.14} \\
\makebox[1.2em][c]{\textbf{\cmark}} & \makebox[1.2em][c]{\textbf{\cmark}} & \makebox[1.2em][c]{\textbf{\cmark}} & \makebox[1.2em][c]{\textbf{\cmark}} & \cellcolor[HTML]{FF9999}\textbf{25.97} & \cellcolor[HTML]{FF9999}\textbf{20.12} & \cellcolor[HTML]{FF9999}\textbf{0.548} & \cellcolor[HTML]{FF9999}\textbf{0.433} & \cellcolor[HTML]{99E4B0}\textbf{24.15} & \cellcolor[HTML]{99E4B0}\textbf{18.92} & \cellcolor[HTML]{d8fcdb}\underline{0.515} & \cellcolor[HTML]{99E4B0}\textbf{0.450} & 22.96 \\
\bottomrule
\end{tabular}%
}
\end{table*}

\subsection{Virtual Exposure Sweeps}
A primary benefit of recovering a true linear HDR radiance field is the ability to post-process rendered views while retaining the scene's original photometric information. To demonstrate this, in \Cref{subfig:exposures} we perform a virtual exposure sweep across three multipliers ($0.50\times$ to $4.00\times$) over both LDR and HDR renderings from our pipeline. While standard LDR renderings suffer severe clipping and detail loss when exposure is scaled, our linear formulation adjusts brightness naturally. Even under a $4.00\times$ boost, our HDR representation preserves structural coherence across shadows and highlights, enabling high-fidelity post-capture photometric editing.

\subsection{Ablation study}

\Cref{tab:ablation} presents an ablation study analyzing the individual and combined contributions of our architectural components across all 10 scenes. When applied individually, the $L_{\text{SSIM}}$ term, the $L_1$-Reinhard loss, and the logarithmic color reparameterization each improve performance over the MonoGS baseline. However, their synergistic combination, coupled with our spatial gradient weighting, is critical for achieving the best performance, driving LDR trajectory error down to $25.97\text{ cm}$ and achieving our best linear HDR trajectory accuracy with an ATE of $24.15\text{ cm}$ and a PSNR-$\mu$ metric of $22.96\text{ dB}$.

\section{Conclusion}
\label{sec:conclusion}

We presented the first online Gaussian SLAM framework that trains and renders directly in 16-bit linear HDR space, bridging the gap between computational photography and online dense visual SLAM. Our HDR GS module integrates seamlessly into existing dense RGB-D Gaussian SLAM and learning-based tracking backbones with minimal modifications, improving tracking robustness under severe illumination variations.
Building upon this, we developed our complete, fully optimized pipeline with the introduction of an HDR-based loss, which substantially improves tracking and robustness while enhancing efficient end-to-end processing. Crucially, our formulation also acts as a regularizer, improving tracking performance when trained on standard LDR images.

While our formulation substantially reduces geometric error over our baseline, reconstruction metrics remain behind methods with stronger image-based reconstruction, suggesting that improved tracking does not necessarily translate to higher perceptual fidelity and that our photometric contributions would benefit from a stronger geometric backbone. Our evaluation is also restricted to indoor sequences with a fixed OptiTrack installation, leaving outdoor scenes and broader camera configurations untested.

We hope this work, with our code and datasets, provides a foundation for HDR-aware online SLAM and active perception research under challenging lighting conditions.

{
    \small
    \bibliographystyle{ieeenat_fullname}
    \bibliography{main}
}

\maketitlesupplementary
\section{Dataset}
\label{sec:suppl_dataset}
Prior visual SLAM benchmarks are restricted to standard, processed LDR frames, completely lacking sequential, unprocessed sensor-linear RAW imagery with ground-truth camera trajectories. To close this gap and enable evaluation under extreme exposure gradients, we introduce RawSLAM.

\subsection{Capture setup}

Data was acquired with an Intel RealSense D435i, which simultaneously streams RAW imagery, depth, and IMU (accelerometer + gyroscope) measurements from a single rigid sensor package. RGB frames were recorded at $30\,\text{fps}$ and $1920{\times}1080$ resolution and saved as 16-bit Bayer RAW \texttt{.dng} files. Depth was captured at $360{\times}640$ resolution and reprojected onto the color viewpoint with
\texttt{rs.align} at capture time, so the released maps are pixel-aligned to
their corresponding RAW frames at $1920{\times}1080$, upsampled from the lower
native depth resolution.

The ten sequences were recorded handheld in the same controlled indoor space at different hours of the day in order to capture different natural lighting conditions, complemented by adjustable artificial lighting, with objects and structures placed within the capture volume to create scenes with distinct illumination and content characteristics: objects that emit light, transparent and reflective surfaces, low-texture regions, etc. The scenes (\Cref{tab:dataset}) deliberately span both very bright, sunlit scenes prone to highlight clipping
(\emph{bottles}, \emph{small\_city}, \emph{nerdy\_robot}) and very dark, low-light scenes dominated by sensor noise (\emph{kitchen}, \emph{candles}), alongside scenes combining strong local light--shadow contrast (\emph{coat\_rack}) and artificial point light sources prone to clipping (\emph{christmas}, \emph{candles}, \emph{cabin}). Sequences range from 1,581 to 2,100 frames, for a total of 18,606 frames.

Ground-truth 6-DoF camera poses were captured with an OptiTrack motion-capture system of eight Flex~13 cameras arranged around the recording volume, operating at $120\,\text{Hz}$ with a nominal marker-tracking accuracy on the order of $0.2\,\text{mm}$ \cite{optitrack}. This external tracking is independent of the visual and inertial streams being evaluated, and avoids the systematic biases of image-based pose estimation (e.g., SfM pipelines such as COLMAP \cite{colmap}) under the very lighting conditions RawSLAM is designed to stress. Reflective markers were rigidly attached directly to the camera in an asymmetric constellation optimized for robust rotational (not just positional) tracking.


\begin{table}[t]
    \centering
    \caption{Overview of our RawSLAM Dataset.}
    \label{tab:dataset}
    \footnotesize
    \renewcommand{\arraystretch}{1.2}
    \resizebox{\linewidth}{!}{%
        \begin{tabular}{l c c l}
            \hline
            \textbf{Scene} & \textbf{Num. of frames} & \textbf{Traj. length (m)} & \textbf{Scene description} \\
            \hline
            bottles & 1937 & 16.09 & Transparent objects \\
            boxes & 2025 & 18.20 & Transparent objects \\
            cabin & 1679 & 17.63 & Low texture \& bright spot light \\
            candles & 2100 & 11.36 & Dark scene \&  candle lights\\
            christmas & 1929 & 22.45 & Artificial little lights \\
            coffee & 1916 & 23.85 & High contrast \&  reflections\\
            coat\_rack & 1911 & 15.66 & High contrast \\
            kitchen & 1805 & 33.20 & Dark scene with backlighting\\
            nerdy\_robot & 1723 & 28.82 & Standard illumination \\
            small\_city & 1581 & 18.82 & Backlighting \\
            \hline
        \end{tabular}
        }
\end{table}
\subsection{Synchronization}
\paragraph{Temporal synchronization.} The RealSense and OptiTrack streams ran on independent clocks and were aligned post hoc. Each recording begins with the camera pointed at a rigid, flat calibration object that is also tracked as its own OptiTrack rigid body: the object is briefly lifted and dropped, producing a sharp, unambiguous motion event visible in both streams. We locate the synchronization instant by plotting the calibration object's tracked height over time and selecting the first frame after the drop at which its vertical position stops changing. Despite the sensor noise this introduces (oscillations of a few tens of millimeters around the settling point), the resulting temporal offset error is small, around $0.0083\,\text{s}$. Both streams' clocks are zeroed at this synchronization instant, and the initial calibration segment preceding it is trimmed from the released sequence.

\paragraph{Frame association.} Because OptiTrack runs at a higher rate ($120\,\text{Hz}$) than the camera ($30\,\text{fps}$), each camera frame is matched to its nearest OptiTrack frame by timestamp after clock alignment. 
Every accepted association is written as one row of \texttt{groundtruth.txt}, with columns \texttt{[frame\_id, timestamp\_ms, x, y, z, rot\_x, rot\_y, rot\_z]} (translation in meters, rotation as XYZ Euler angles in degrees, both relative to the post-synchronization clock). IMU samples are stored individually per frame as a 7-dimensional vector containing a relative timestamp offset together with the 3-axis accelerometer and 3-axis gyroscope readings captured within that frame's exposure window.

\subsection{Release format}
Each scene is released as a self-contained folder:
\begin{itemize}
    \item \texttt{raw/} -- unprocessed 16-bit Bayer RAW frames (\texttt{.dng}).
    \item \texttt{depth/} -- depth maps aligned to the RAW/RGB frame.
    \item \texttt{imu/} -- one \texttt{.npy} IMU sample per frame.
    \item \texttt{groundtruth.txt} -- synchronized OptiTrack poses, as described above.
    \item \texttt{my\_dataset.yaml} -- camera intrinsics and depth scale (provided by the camera).
\end{itemize}
We distribute only the raw sensor data (RAW, depth, IMU, poses) rather than the decoded image streams: both the 16-bit linear HDR and the 8-bit tonemapped LDR streams used throughout the paper are derived from the same RAW frames by our open-sourced \texttt{extract.py} script (\Cref{sec:raw2hdr}), so both streams can be regenerated bit-exactly and future work can experiment with alternative demosaicing or tonemapping choices from the same raw captures. Our capture toolkit (recording and temporal synchronization) will be released as a separate repository alongside the dataset.


For the models we tested (MonoGS, SplaTAM, Gaussian SLAM, DROID-W, and ORB-SLAM2), we also provide data loaders based on the TUM RGB-D \cite{tum} parsers, given their common use in SLAM research and similar RGB-D indoor-scene structure.

\section{Per-scene results}

The per-scene results in \Cref{tab:results_ate_per_scene,tab:results_rend_per_scene}
show that the benefit of native HDR depends on scene illumination. Our HDR configuration improves tracking over our LDR configuration on six of the ten sequences, with the largest gains in ATE RMSE on \textit{boxes} ($31.53 \rightarrow 15.19$ cm) and \textit{candles} ($23.59 \rightarrow 15.58$ cm), scenes containing transparency, very dark regions, or artificial bright light sources, making them particularly sensitive to highlight clipping and shadow quantization.

These gains are not due to HDR input alone. The per-scene ablation in \Cref{tab:ablation_per_scene} shows that our complete formulation reduces ATE over Vanilla HDR on nine of the ten scenes, most notably \textit{nerdy\_robot} ($59.85 \rightarrow 6.69$ cm), \textit{candles} ($57.98 \rightarrow 15.58$ cm), \textit{bottles} ($61.20 \rightarrow 27.91$ cm), \textit{cabin} ($42.54 \rightarrow 9.97$ cm) and \textit{coffee} ($51.37 \rightarrow 23.15$ cm), with \textit{coat\_rack} as the only exception. The best HDR trajectory, $6.69$ cm on \textit{nerdy\_robot}, a standard-illumination scene, suggests that the formulation also acts as a regularizer beyond extreme HDR conditions.

We can observe in \Cref{tab:results_rend_per_scene} that in rendering metrics MonoGS (both the only-HDR-module and our whole configuration versions) obtain the highest PSNR-$\mu$ on every one of the ten scenes among all other evaluated methods using our HDR module, and reaches its own strongest radiance fidelity on \textit{nerdy\_robot} ($26.71$ dB). 

However, HDR does not win everywhere: ATE RMSE regresses in some scenes when trained on HDR with respect to LDR, as in \textit{cabin} ($6.40 \rightarrow 9.97$ cm), \textit{kitchen} ($16.07 \rightarrow 17.14$ cm), and especially in \textit{coat\_rack} ($61.24 \rightarrow 71.18$ cm). 

Overall, HDR is most useful when informative structure lies across severely under- and overexposed regions, while the proposed optimization is what converts that information into stable online tracking.

\section{MLP-based color parameterization}

To validate our MLP-free logarithmic parameterization approach, we tested an adaptation of the offline raw reconstruction framework of LE3D~\cite{le3d} to our online SLAM loop, using a lightweight three-layer color Multi-Layer Perceptron (MLP) with exponential activation to parameterize the linear HDR color space. We jointly optimize the MLP parameters along with camera poses and Gaussian attributes during tracking and mapping. The results we obtained, evaluated on 8 scenes (all but \textit{kitchen} and \textit{coat\_rack}), are collected in \Cref{tab:mlp}.

\begin{table}[t]
\centering
\caption{Quantitative results across eight of the RawSLAM scenes and three runs. Vanilla HDR stands for the HDR adaptation of MonoGS without any of our further changes, while Full pipeline includes all our modifications. Rendering metrics were computed over the tonemapped version of the HDR renders.}
\label{tab:mlp}
\renewcommand{\arraystretch}{1.2}
\resizebox{\linewidth}{!}{%
\begin{tabular}{lccccc}
\toprule
 & \begin{tabular}{@{}c@{}} \textbf{ATE}  (cm) $\downarrow$\end{tabular} 
 & \begin{tabular}{@{}c@{}}\textbf{PSNR}  (dB) $\uparrow$\end{tabular} 
 & \begin{tabular}{@{}c@{}}\textbf{SSIM}  $\uparrow$\end{tabular} 
 & \begin{tabular}{@{}c@{}}\textbf{LPIPS}  $\downarrow$\end{tabular} 
 & \begin{tabular}{@{}c@{}}\textbf{FPS}  $\uparrow$\end{tabular} 
 \\ \midrule
LDR  & 58.79  & \cellcolor[HTML]{fffbda}18.40 & 0.501 & 0.570 & \cellcolor[HTML]{fffbda} 1.11 \\
\cdashline{1-6}
Vanilla HDR & 48.75 & 17.63 & \cellcolor[HTML]{d8fcdb}\underline{0.579} & 0.553 & 1.08\\ 
+ MLP & \cellcolor[HTML]{fffbda}41.47 & \cellcolor[HTML]{fffbda}18.40 & \cellcolor[HTML]{99E4B0}\textbf{0.618} & \cellcolor[HTML]{d8fcdb}\underline{0.468} & 0.91 \\
+ HDR GS Module &  \cellcolor[HTML]{d8fcdb}\underline{36.37} & \cellcolor[HTML]{d8fcdb}\underline{18.82} & 0.515 & \cellcolor[HTML]{fffbda}0.469 & \cellcolor[HTML]{d8fcdb}\underline{1.13} \\
Full pipeline & \cellcolor[HTML]{99E4B0}\textbf{19.15} & \cellcolor[HTML]{99E4B0}\textbf{19.07} & \cellcolor[HTML]{fffbda}0.537  & \cellcolor[HTML]{99E4B0}\textbf{0.436} & \cellcolor[HTML]{99E4B0}\textbf{1.22} \\
\bottomrule
\end{tabular}%
}
\end{table}

The MLP obtains higher SSIM ($0.618$ vs.\ $0.515$) but lower PSNR ($18.40$ vs.\ $18.82$ dB) respect to the use of the HDR GS Module. 

Moreover, the MLP improves tracking over Vanilla HDR ($48.75 \rightarrow 41.47$ cm), but our direct parameterization, HDR GS Module, leads to better results ($36.37$ cm). Our full pipeline reaches $19.15$ cm, but HDR GS Module is what the MLP approach results should be compared to, as they do not contain any of our other changes. 

Along with the tracking score, we chose our MLP-free formulation because of the slower running time of the MLP approach. Evaluating the network per primitive and backpropagating through it during both rendering and optimization results in a processing speed of $0.91$ FPS, whereas the logarithmic parameterization requires only element-wise operations, achieving an optimization speed of $1.13$ FPS.

\section{Implementation details}
All hyperparameters, including input resolution and downsampling factor, are
released as configuration files, alongside environment specifications and
Dockerfiles recording the exact software environment used.

\subsection{Raw to HDR/LDR pipeline}
\label{sec:raw2hdr}
Our provided file \texttt{extract.py} unzips the downloaded scene and converts each RAW frame referenced in \texttt{groundtruth.txt} into two aligned streams using \texttt{rawpy}, sharing the demosaicing, camera white balance, and camera-to-sRGB color correction matrix (stages 1--2 of \Cref{fig:image_processing_pipeline}). The two streams differ {only} in the tonemapping and quantization step, isolating precisely the variable studied throughout the paper.

\begin{itemize}
    \item \textbf{RAW $\rightarrow$ HDR.} We disable the tonemapping curve (gamma $=(1.0,\,0.0)$) and automatic brightness, and decode to 16 bits per channel. This yields a scene-linear image in sRGB primaries that we treat as our HDR stream: pixel values are proportional to sensor irradiance and are only bounded by the sensor's native dynamic range, not by an 8-bit display encoding. 

    \item \textbf{RAW $\rightarrow$ LDR.} We apply the standard sRGB gamma curve (gamma $=(2.2,\,0.0)$) and decode to 8 bits per channel, reproducing what a conventional camera ISP would output (stage 3 of \Cref{fig:image_processing_pipeline}) and serving as the LDR counterpart used by our baseline comparisons.
        
\end{itemize}

After extraction, images are saved in \texttt{.png} format and we automatically verify that every frame referenced in \texttt{groundtruth.txt} was successfully decoded and that the HDR/LDR streams have the expected 16/8 bit depth, flagging any missing or malformed frame before training.

\subsection{Evaluation details}
\label{sec:evaldetails}

\begin{figure*}[t]
  \centering
  \small
  \setlength{\tabcolsep}{2pt}
  
  \begin{tabular}{c c c c c}
    & \textbf{Baseline} & \textbf{Ours (LDR)} & \textbf{Ours (HDR)} & \textbf{Ground-truth} \\[6pt]
        

    \rowlabel{boxes}{2025} &
    \gridimg{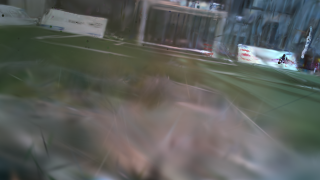} &
    \gridimg{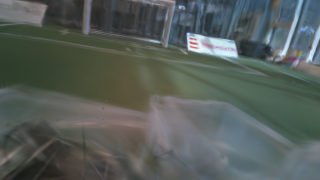} &
    \gridimg{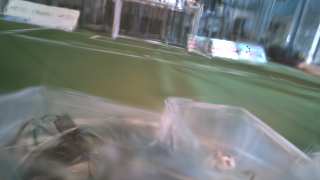} &
    \gridimg{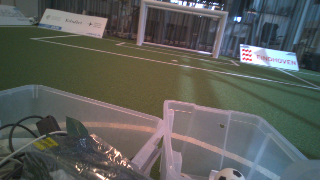} \\[30pt]
    
    \rowlabel{cabin}{1679} &
    \gridimg{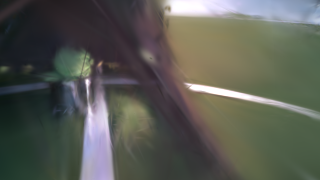} &
    \gridimg{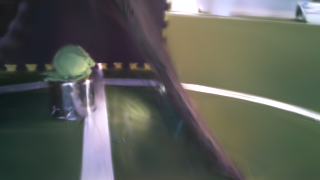} &
    \gridimg{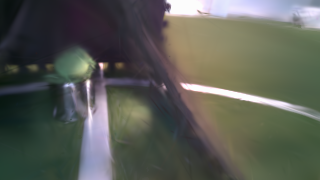} &
    \gridimg{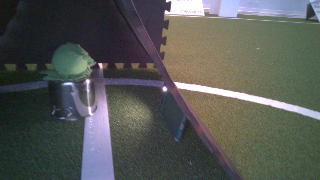} \\[30pt]
    
    \rowlabel{candles}{2100} &
    \gridimg{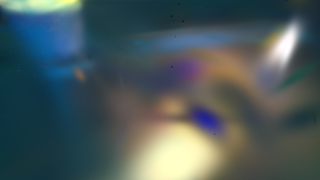} &
    \gridimg{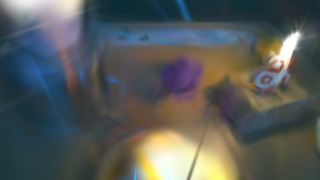} &
    \gridimg{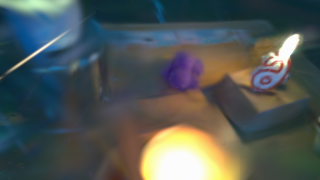} &
    \gridimg{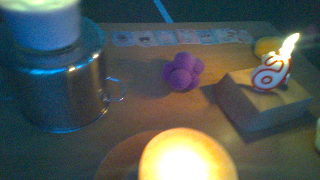} \\[30pt]

    \rowlabel{christmas}{1929} &
    \gridimg{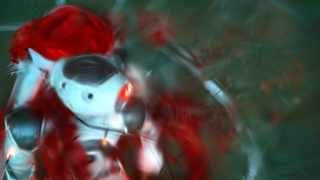} &
    \gridimg{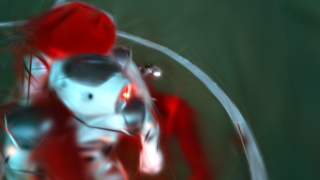} &
    \gridimg{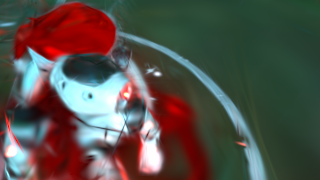} &
    \gridimg{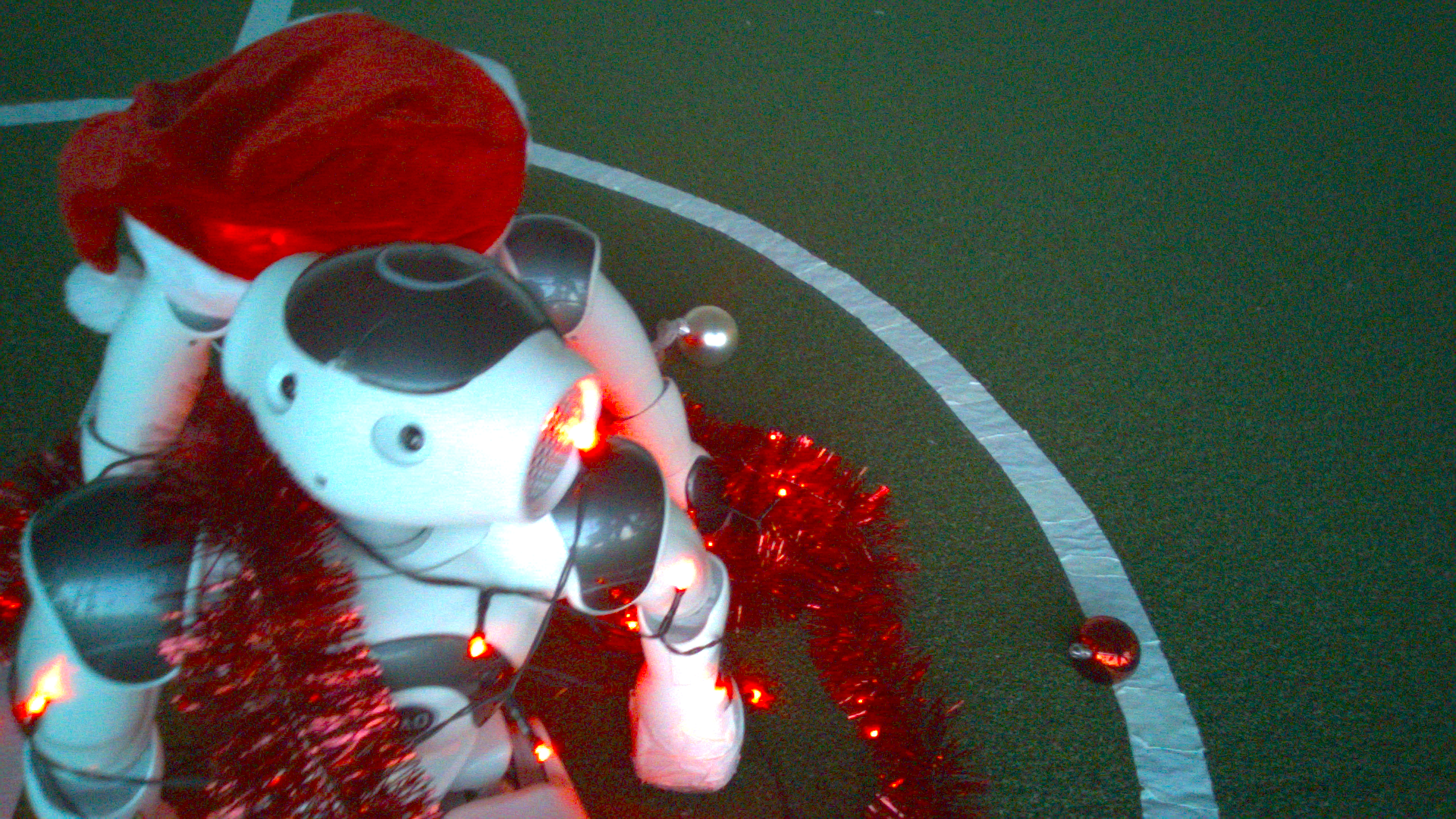} \\[30pt]

    
    
    \rowlabel{kitchen}{1805} &
    \gridimg{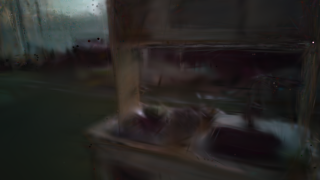} &
    \gridimg{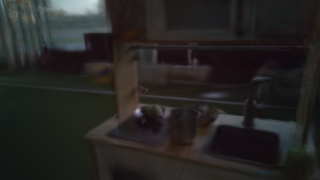} &
    \gridimg{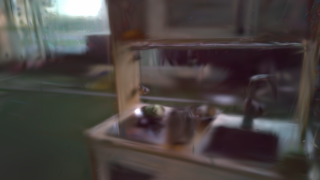} &
    \gridimg{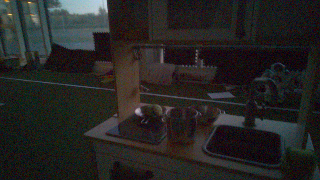} \\[30pt]
    
    \rowlabel{nerdy\_robot}{1723} &
    \gridimg{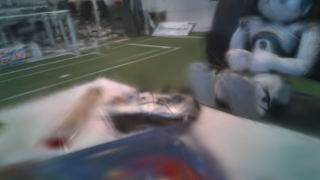} &
    \gridimg{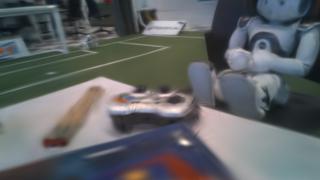} &
    \gridimg{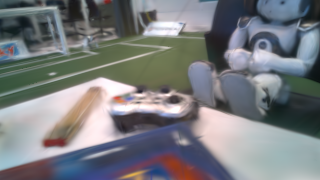} &
    \gridimg{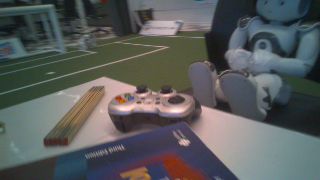} \\[30pt]
    
    \rowlabel{small\_city}{1581} &
    \gridimg{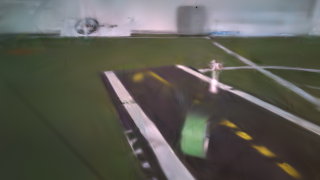} &
    \gridimg{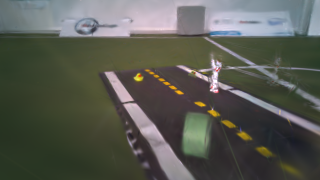} &
    \gridimg{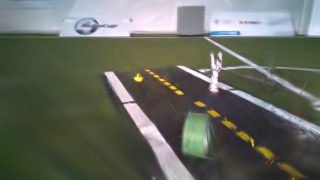} &
    \gridimg{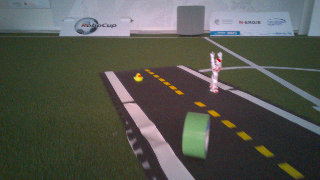} \\
  \end{tabular}

  \caption{Rendering performance on RawSLAM dataset scenes that were not shown in \Cref{fig:three_renders}.}
  \label{fig:rendering}
\end{figure*}

\paragraph{Tonemapped metrics.} PSNR, SSIM and LPIPS are defined for LDR images and cannot be applied to HDR images directly. Therefore, we perform an HDR to LDR transformation to convert our obtained HDR renders to LDR format. For that, we rescale each image using its percentile intensity as the white level and we apply the sRGB gamma curve to it. The same routine is used for producing the display-referred images in
\Cref{fig:main,fig:rendering,fig:three_renders}.

\paragraph{PSNR-$\mu$.} We also implemented the computation of the PSNR-$\mu$~\cite{pu-psnr} metric for measuring the HDR rendering quality. This metric is obtained by computing the PSNR between the compressed linear radiance of the rendered and ground-truth images ($T_\mu(I_R)$ and $T_\mu(I_{GT})$) in the following way:
\begin{equation}
    T_\mu(x) = \frac{\log(1 + \mu x)}{\log(1 + \mu)},
    \label{eq:mulaw}
\end{equation}

\noindent where $\mu = 5000$. As $T_\mu$ is not scale invariant, we apply no per-frame or per-scene rescaling before compression: both inputs enter \Cref{eq:mulaw} on the fixed scale set at data loading, where 16-bit frames are divided by $65535$ and clamped to $[0,1]$, identically for every scene and method. This normalization is global, so PSNR-$\mu$ is comparable across scenes as well as methods. 

\paragraph{Failure criterion.} A run is considered to have failed if the tracking front-end crashes or
terminates early, or if the final ATE RMSE exceeds $2.5\,\text{m}$, a
magnitude at which, for trajectories of $11$--$33\,\text{m}$ inside a single
room-sized volume, the estimate has lost correspondence with the scene.
Failures are counted per run, and a scene is \red{FAIL} only when all three runs fail. Means are computed over the surviving runs.

\paragraph{Render saving.} We extend the host models' render-saving utilities
to emit 16-bit outputs, as their default writers quantize to 8 bits and would
discard the dynamic range the pipeline recovers.
\subsection{Plug-and-play integration}
\label{sec:plyandplayintegration}

Besides inserting the HDR GS module, enabling linear HDR processing in each host
framework required only a small and structurally identical set of changes. The
image loading, normalization and saving paths were the one unavoidable
exception: several standard libraries (e.g., Pillow) offer limited support for
16-bit RGB images, so these were reimplemented with OpenCV.

Each dataset loader was extended with a \texttt{RawSLAM} variant that reads
16-bit linear frames, validates the decoded array against \texttt{dtype ==
uint16}, and normalizes by $65535$ rather than $255$. It is added alongside the
existing 8-bit path, so both regimes remain selectable through a single
\texttt{raw} configuration flag.

Our rasterizer exposes $\alpha_\mathrm{threshold}$ as a run-time field rather
than the hardcoded $1/255$ constant, so every render call must supply it
explicitly. Each pipeline precomputes it as a binary choice, 
$\alpha_\mathrm{threshold} \in \{1/255,\ 1/65535\}$, selected by the same
\texttt{raw} flag, and passes it at every call site: initialization, tracking,
mapping, densification, and evaluation.


\begin{table*}[t]
\centering
\caption{ATE RMSE (cm, $\downarrow$) per scene, averaged across runs only. The \colorbox[HTML]{99E4B0}{\textbf{first}}, \colorbox[HTML]{d8fcdb}{\underline{second}} and \colorbox[HTML]{fffbda}{third} best results are highlighted across both HDR and LDR, while the results from DROID-W and ORB-SLAM2 are excluded from the ranking as they should be seen as \textbf{external feature-based tracking references rather than direct photometric 3DGS baselines.}}
\label{tab:results_ate_per_scene}
\footnotesize
\resizebox{\linewidth}{!}{%
\begin{tabular}{lccccccccccc}
\toprule
Method & bottles & boxes & cabin & candles & christmas & coat rack & coffee & kitchen & nerdy robot & small city & \textbf{Mean} \\ \midrule
Droid-W & 10.46 & 9.16 & 7.34 & 7.37 & 7.72 & 71.42 & \red{FAIL} & 6.77 & 8.17 & 18.27 & 16.30 \\
ORB-SLAM 2 & 3.51 & 2.05 & 0.89 & 0.75 & 1.57 & 7.37 & 18.01 & 0.42 & 4.38 & 1.07 & 4.00 \\
\cdashline{1-12}
\rule{0pt}{2.2ex} MonoGS \cite{monogs} & 52.61 & 48.30 & \cellcolor[HTML]{fffbda}23.56 & 74.22 & 38.76 & \cellcolor[HTML]{99E4B0}\textbf{60.99} & 74.85 & 37.06 & 68.39 & 89.61 & 56.84 \\
SplaTAM \cite{splatam} & 85.42 & 163.02 & 48.76 & \red{FAIL} & 57.89 & 220.88 & \red{FAIL} & \cellcolor[HTML]{fffbda}23.24 & \cellcolor[HTML]{fffbda}25.84 & \red{FAIL} & 89.29 \\
Gaussian SLAM \cite{gaussianslam} & \cellcolor[HTML]{99E4B0}\textbf{13.88} & 79.98 & 96.04 & \red{FAIL} & 177.55 & 75.13 & 126.19 & 31.74 & 39.90 & 141.96 & 86.93 \\
Ours (LDR) & \cellcolor[HTML]{d8fcdb}\underline{27.00} & \cellcolor[HTML]{d8fcdb}\underline{31.53} & \cellcolor[HTML]{99E4B0}\textbf{6.40} & \cellcolor[HTML]{d8fcdb}\underline{23.59} & \cellcolor[HTML]{d8fcdb}\underline{29.89} & \cellcolor[HTML]{d8fcdb}\underline{61.24} & \cellcolor[HTML]{d8fcdb}\underline{26.83} & \cellcolor[HTML]{99E4B0}\textbf{16.07} & \cellcolor[HTML]{d8fcdb}\underline{7.36} & \cellcolor[HTML]{d8fcdb}\underline{29.82} & \cellcolor[HTML]{d8fcdb}\underline{25.97} \\
\cmidrule(l){1-12}
Vanilla HDR & 61.20 & \cellcolor[HTML]{fffbda}34.06 & 42.54 & \cellcolor[HTML]{fffbda}57.98 & \cellcolor[HTML]{fffbda}36.75 & \cellcolor[HTML]{fffbda}63.71 & \cellcolor[HTML]{fffbda}51.37 & 40.58 & 59.85 & \cellcolor[HTML]{fffbda}46.25 & \cellcolor[HTML]{fffbda}49.43 \\
Ours (HDR) & \cellcolor[HTML]{fffbda}27.91 & \cellcolor[HTML]{99E4B0}\textbf{15.19} & \cellcolor[HTML]{d8fcdb}\underline{9.97} & \cellcolor[HTML]{99E4B0}\textbf{15.58} & \cellcolor[HTML]{99E4B0}\textbf{25.70} & 71.18 & \cellcolor[HTML]{99E4B0}\textbf{23.15} & \cellcolor[HTML]{d8fcdb}\underline{17.14} & \cellcolor[HTML]{99E4B0}\textbf{6.69} & \cellcolor[HTML]{99E4B0}\textbf{29.02} & \cellcolor[HTML]{99E4B0}\textbf{24.15} \\
\bottomrule
\end{tabular}%
}
\end{table*}
\begin{table*}[t]
\centering
\caption{Rendering metrics (PSNR, SSIM, LPIPS, Depth $L_1$ (cm), and PSNR-$\mu$ for the HDR pipeline) per scene, averaged over runs. Average over scenes is also included in the last column. \protect\footnotemark}
\label{tab:results_rend_per_scene}
\footnotesize
\resizebox{\linewidth}{!}{%
\begin{tabular}{llccccccccccc}
\toprule
Method & Metric & bottles & boxes & cabin & candles & christmas & coat rack & coffee & kitchen & nerdy robot & small city & \textbf{Mean} \\ \midrule
\multirow{4}{*}{DROID-W \cite{droidw}} & PSNR $\uparrow$ & 17.07 & 19.41 & 19.91 & 16.60 & 17.34 & 16.50 & \red{FAIL} & 21.87 & 18.64 & 19.73 & 18.56 \\
 & SSIM $\uparrow$ & 0.560 & 0.661 & 0.622 & 0.450 & 0.370 & 0.555 & \red{FAIL} & 0.370 & 0.645 & 0.567 & 0.533 \\
 & LPIPS $\downarrow$ & 0.532 & 0.478 & 0.525 & 0.587 & 0.564 & 0.540 & \red{FAIL} & 0.554 & 0.516 & 0.463 & 0.529 \\
 & Depth $L_1$ (cm) $\downarrow$ & 28.92 & 21.55 & 23.01 & 12.70 & 22.37 & 35.07 & \red{FAIL} & 34.35 & 18.65 & 30.46 & 25.23 \\
\cdashline{1-13}
\rule{0pt}{2.2ex} \multirow{4}{*}{SplaTAM \cite{splatam}} & PSNR $\uparrow$ & \cellcolor[HTML]{fffbda}17.67 & 19.66 & 19.83 & \red{FAIL} & \cellcolor[HTML]{fffbda}18.66 & 15.38 & \red{FAIL} & \cellcolor[HTML]{FF9999}\textbf{22.25} & \cellcolor[HTML]{fffbda}20.45 & \red{FAIL} & \cellcolor[HTML]{fffbda}19.13 \\
 & SSIM $\uparrow$ & \cellcolor[HTML]{FCDFD8}\underline{0.675} & \cellcolor[HTML]{FCDFD8}\underline{0.748} & \cellcolor[HTML]{FCDFD8}\underline{0.748} & \red{FAIL} & \cellcolor[HTML]{FCDFD8}\underline{0.649} & \cellcolor[HTML]{FCDFD8}\underline{0.593} & \red{FAIL} & \cellcolor[HTML]{FCDFD8}\underline{0.720} & \cellcolor[HTML]{FCDFD8}\underline{0.812} & \red{FAIL} & \cellcolor[HTML]{FCDFD8}\underline{0.706} \\
 & LPIPS $\downarrow$ & \cellcolor[HTML]{FCDFD8}\underline{0.399} & \cellcolor[HTML]{FCDFD8}\underline{0.328} & \cellcolor[HTML]{FCDFD8}\underline{0.396} & \red{FAIL} & \cellcolor[HTML]{FCDFD8}\underline{0.470} & \cellcolor[HTML]{fffbda}0.477 & \red{FAIL} & \cellcolor[HTML]{FCDFD8}\underline{0.511} & \cellcolor[HTML]{FCDFD8}\underline{0.257} & \red{FAIL} & \cellcolor[HTML]{FCDFD8}\underline{0.405} \\
 & Depth $L_1$ (cm) $\downarrow$ & \cellcolor[HTML]{FCDFD8}\underline{142.97} & \cellcolor[HTML]{FCDFD8}\underline{89.24} & \cellcolor[HTML]{FCDFD8}\underline{42.90} & \red{FAIL} & \cellcolor[HTML]{FCDFD8}\underline{24.88} & \cellcolor[HTML]{fffbda}192.16 & \red{FAIL} & \cellcolor[HTML]{fffbda}198.72 & \cellcolor[HTML]{FCDFD8}\underline{42.91} & \red{FAIL} & \cellcolor[HTML]{FCDFD8}\underline{104.83} \\
\cdashline{1-13}
\rule{0pt}{2.2ex} \multirow{4}{*}{Gaussian SLAM \cite{gaussianslam}} & PSNR $\uparrow$ & \cellcolor[HTML]{FF9999}\textbf{20.71} & \cellcolor[HTML]{FF9999}\textbf{23.54} & \cellcolor[HTML]{FF9999}\textbf{23.35} & \red{FAIL} & \cellcolor[HTML]{FF9999}\textbf{21.34} & \cellcolor[HTML]{FF9999}\textbf{20.80} & \cellcolor[HTML]{FF9999}\textbf{21.16} & \cellcolor[HTML]{fffbda}21.45 & \cellcolor[HTML]{FF9999}\textbf{23.32} & \cellcolor[HTML]{FF9999}\textbf{21.10} & \cellcolor[HTML]{FF9999}\textbf{21.86} \\
 & SSIM $\uparrow$ & \cellcolor[HTML]{FF9999}\textbf{0.801} & \cellcolor[HTML]{FF9999}\textbf{0.847} & \cellcolor[HTML]{FF9999}\textbf{0.835} & \red{FAIL} & \cellcolor[HTML]{FF9999}\textbf{0.777} & \cellcolor[HTML]{FF9999}\textbf{0.828} & \cellcolor[HTML]{FF9999}\textbf{0.791} & \cellcolor[HTML]{FF9999}\textbf{0.729} & \cellcolor[HTML]{FF9999}\textbf{0.866} & \cellcolor[HTML]{FF9999}\textbf{0.755} & \cellcolor[HTML]{FF9999}\textbf{0.803} \\
 & LPIPS $\downarrow$ & \cellcolor[HTML]{FF9999}\textbf{0.360} & \cellcolor[HTML]{FF9999}\textbf{0.320} & \cellcolor[HTML]{fffbda}0.408 & \red{FAIL} & \cellcolor[HTML]{FF9999}\textbf{0.388} & \cellcolor[HTML]{FF9999}\textbf{0.335} & \cellcolor[HTML]{FF9999}\textbf{0.407} & \cellcolor[HTML]{FF9999}\textbf{0.479} & \cellcolor[HTML]{fffbda}0.302 & \cellcolor[HTML]{FF9999}\textbf{0.452} & \cellcolor[HTML]{FF9999}\textbf{0.384} \\
 & Depth $L_1$ (cm) $\downarrow$ & \cellcolor[HTML]{FF9999}\textbf{66.41} & \cellcolor[HTML]{FF9999}\textbf{39.83} & \cellcolor[HTML]{FF9999}\textbf{27.56} & \red{FAIL} & \cellcolor[HTML]{FF9999}\textbf{15.98} & \cellcolor[HTML]{FF9999}\textbf{87.58} & \cellcolor[HTML]{FF9999}\textbf{50.97} & \cellcolor[HTML]{FF9999}\textbf{94.22} & \cellcolor[HTML]{FF9999}\textbf{34.30} & \cellcolor[HTML]{FF9999}\textbf{82.33} & \cellcolor[HTML]{FF9999}\textbf{55.46} \\
\cdashline{1-13}
\rule{0pt}{2.2ex} \multirow{4}{*}{MonoGS \cite{monogs}} & PSNR $\uparrow$ & 16.37 & \cellcolor[HTML]{fffbda}20.06 & \cellcolor[HTML]{fffbda}19.88 & \cellcolor[HTML]{FCDFD8}\underline{18.50} & \cellcolor[HTML]{fffbda}18.66 & \cellcolor[HTML]{fffbda}16.26 & \cellcolor[HTML]{fffbda}16.89 & 20.17 & 18.36 & \cellcolor[HTML]{fffbda}18.46 & 18.36 \\
 & SSIM $\uparrow$ & 0.464 & 0.592 & 0.586 & \cellcolor[HTML]{FCDFD8}\underline{0.436} & 0.398 & 0.496 & \cellcolor[HTML]{fffbda}0.450 & 0.324 & 0.589 & \cellcolor[HTML]{fffbda}0.494 & 0.483 \\
 & LPIPS $\downarrow$ & 0.639 & 0.477 & 0.540 & \cellcolor[HTML]{FCDFD8}\underline{0.599} & 0.527 & 0.485 & \cellcolor[HTML]{fffbda}0.670 & 0.645 & 0.510 & \cellcolor[HTML]{fffbda}0.598 & 0.569 \\
 & Depth $L_1$ (cm) $\downarrow$ & 236.97 & 169.66 & 76.85 & \cellcolor[HTML]{FCDFD8}\underline{81.26} & 68.51 & 212.84 & \cellcolor[HTML]{fffbda}170.10 & 217.30 & 117.51 & \cellcolor[HTML]{fffbda}196.18 & 154.72 \\
\cdashline{1-13}
\rule{0pt}{2.2ex} \multirow{4}{*}{Ours (LDR)} & PSNR $\uparrow$ & \cellcolor[HTML]{FCDFD8}\underline{17.87} & \cellcolor[HTML]{FCDFD8}\underline{21.66} & \cellcolor[HTML]{FCDFD8}\underline{22.18} & \cellcolor[HTML]{FF9999}\textbf{20.57} & \cellcolor[HTML]{FCDFD8}\underline{19.00} & \cellcolor[HTML]{FCDFD8}\underline{17.33} & \cellcolor[HTML]{FCDFD8}\underline{18.77} & \cellcolor[HTML]{FCDFD8}\underline{21.82} & \cellcolor[HTML]{FCDFD8}\underline{22.59} & \cellcolor[HTML]{FCDFD8}\underline{19.39} & \cellcolor[HTML]{FCDFD8}\underline{20.12} \\
 & SSIM $\uparrow$ & \cellcolor[HTML]{fffbda}0.537 & \cellcolor[HTML]{fffbda}0.665 & \cellcolor[HTML]{fffbda}0.638 & \cellcolor[HTML]{FF9999}\textbf{0.497} & \cellcolor[HTML]{fffbda}0.424 & \cellcolor[HTML]{fffbda}0.549 & \cellcolor[HTML]{FCDFD8}\underline{0.517} & \cellcolor[HTML]{fffbda}0.367 & \cellcolor[HTML]{fffbda}0.758 & \cellcolor[HTML]{FCDFD8}\underline{0.526} & \cellcolor[HTML]{fffbda}0.548 \\
 & LPIPS $\downarrow$ & \cellcolor[HTML]{fffbda}0.456 & \cellcolor[HTML]{fffbda}0.351 & \cellcolor[HTML]{FF9999}\textbf{0.392} & \cellcolor[HTML]{FF9999}\textbf{0.446} & \cellcolor[HTML]{fffbda}0.483 & \cellcolor[HTML]{FCDFD8}\underline{0.452} & \cellcolor[HTML]{FCDFD8}\underline{0.532} & \cellcolor[HTML]{fffbda}0.517 & \cellcolor[HTML]{FF9999}\textbf{0.226} & \cellcolor[HTML]{FCDFD8}\underline{0.469} & \cellcolor[HTML]{fffbda}0.433 \\
 & Depth $L_1$ (cm) $\downarrow$ & \cellcolor[HTML]{fffbda}143.75 & \cellcolor[HTML]{fffbda}119.90 & \cellcolor[HTML]{fffbda}50.47 & \cellcolor[HTML]{FF9999}\textbf{51.83} & \cellcolor[HTML]{fffbda}45.38 & \cellcolor[HTML]{FCDFD8}\underline{177.05} & \cellcolor[HTML]{FCDFD8}\underline{101.48} & \cellcolor[HTML]{FCDFD8}\underline{172.97} & \cellcolor[HTML]{fffbda}58.31 & \cellcolor[HTML]{FCDFD8}\underline{152.05} & \cellcolor[HTML]{fffbda}107.32 \\
\cmidrule(l){1-13}
\rule{0pt}{2.2ex} \multirow{5}{*}{DROID-W (HDR)} & PSNR $\uparrow$ & 15.57 & 17.18 & 17.35 & 13.29 & 15.07 & 14.84 & 14.11 & 18.46 & 16.11 & 17.88 & 15.99 \\
 & SSIM $\uparrow$ & 0.458 & 0.536 & 0.512 & 0.340 & 0.294 & 0.437 & 0.386 & 0.259 & 0.519 & 0.469 & 0.421 \\
 & LPIPS $\downarrow$ & 0.574 & 0.544 & 0.636 & 0.636 & 0.574 & 0.622 & 0.767 & 0.655 & 0.551 & 0.564 & 0.612 \\
 & Depth $L_1$ (cm) $\downarrow$ & 31.51 & 19.93 & 27.86 & 13.23 & 21.06 & 33.68 & 40.39 & 28.09 & 19.40 & 30.60 & 26.57 \\
 & PSNR-$\mu$ $\uparrow$ & 19.24 & 22.60 & 20.08 & 18.44 & 17.89 & 17.78 & 16.93 & 18.74 & 20.84 & 21.74 & 19.43 \\
\cdashline{1-13}
\rule{0pt}{2.2ex} \multirow{5}{*}{SplaTAM (HDR) } & PSNR $\uparrow$ & 14.18 & 17.17 & 16.36 & 17.72 & 14.78 & 14.44 & 16.68 & \cellcolor[HTML]{99E4B0}\textbf{20.80} & 17.69 & \cellcolor[HTML]{fffbda}17.62 & 16.74 \\
 & SSIM $\uparrow$ & \cellcolor[HTML]{fffbda}0.470 & \cellcolor[HTML]{d8fcdb}\underline{0.617} & 0.553 & \cellcolor[HTML]{d8fcdb}\underline{0.622} & \cellcolor[HTML]{d8fcdb}\underline{0.438} & \cellcolor[HTML]{fffbda}0.531 & \cellcolor[HTML]{d8fcdb}\underline{0.598} & \cellcolor[HTML]{99E4B0}\textbf{0.620} & \cellcolor[HTML]{fffbda}0.678 & \cellcolor[HTML]{d8fcdb}\underline{0.681} & \cellcolor[HTML]{d8fcdb}\underline{0.581} \\
 & LPIPS $\downarrow$ & 0.562 & 0.453 & 0.601 & \cellcolor[HTML]{fffbda}0.481 & 0.677 & 0.535 & \cellcolor[HTML]{99E4B0}\textbf{0.491} & \cellcolor[HTML]{99E4B0}\textbf{0.616} & \cellcolor[HTML]{fffbda}0.399 & \cellcolor[HTML]{99E4B0}\textbf{0.372} & \cellcolor[HTML]{fffbda}0.519 \\
 & Depth $L_1$ (cm) $\downarrow$ & \cellcolor[HTML]{d8fcdb}\underline{133.89} & \cellcolor[HTML]{d8fcdb}\underline{64.96} & \cellcolor[HTML]{d8fcdb}\underline{49.33} & \cellcolor[HTML]{fffbda}61.21 & \cellcolor[HTML]{d8fcdb}\underline{26.82} & \cellcolor[HTML]{d8fcdb}\underline{175.79} & \cellcolor[HTML]{d8fcdb}\underline{88.63} & \cellcolor[HTML]{fffbda}185.06 & \cellcolor[HTML]{d8fcdb}\underline{44.78} & \cellcolor[HTML]{d8fcdb}\underline{157.43} & \cellcolor[HTML]{d8fcdb}\underline{98.79} \\
 & PSNR-$\mu$ $\uparrow$ & 18.97 & 22.45 & 20.97 & \cellcolor[HTML]{fffbda}17.57 & 15.60 & 17.93 & 20.20 & \cellcolor[HTML]{fffbda}16.05 & 21.76 & 19.70 & 19.12 \\
\cdashline{1-13}
\rule{0pt}{2.2ex} \multirow{5}{*}{Gaussian SLAM (HDR)} & PSNR $\uparrow$ & \cellcolor[HTML]{99E4B0}\textbf{19.21} & \cellcolor[HTML]{99E4B0}\textbf{20.46} & \cellcolor[HTML]{99E4B0}\textbf{21.90} & \cellcolor[HTML]{fffbda}18.38 & \cellcolor[HTML]{99E4B0}\textbf{20.38} & \cellcolor[HTML]{99E4B0}\textbf{19.47} & \cellcolor[HTML]{99E4B0}\textbf{19.95} & \cellcolor[HTML]{fffbda}17.88 & \cellcolor[HTML]{d8fcdb}\underline{20.86} & \cellcolor[HTML]{99E4B0}\textbf{19.33} & \cellcolor[HTML]{99E4B0}\textbf{19.78} \\
 & SSIM $\uparrow$ & \cellcolor[HTML]{99E4B0}\textbf{0.727} & \cellcolor[HTML]{99E4B0}\textbf{0.751} & \cellcolor[HTML]{99E4B0}\textbf{0.782} & \cellcolor[HTML]{99E4B0}\textbf{0.651} & \cellcolor[HTML]{99E4B0}\textbf{0.738} & \cellcolor[HTML]{99E4B0}\textbf{0.761} & \cellcolor[HTML]{99E4B0}\textbf{0.719} & \cellcolor[HTML]{d8fcdb}\underline{0.593} & \cellcolor[HTML]{99E4B0}\textbf{0.800} & \cellcolor[HTML]{99E4B0}\textbf{0.682} & \cellcolor[HTML]{99E4B0}\textbf{0.720} \\
 & LPIPS $\downarrow$ & \cellcolor[HTML]{d8fcdb}\underline{0.466} & \cellcolor[HTML]{fffbda}0.447 & \cellcolor[HTML]{d8fcdb}\underline{0.523} & 0.650 & \cellcolor[HTML]{fffbda}0.516 & \cellcolor[HTML]{d8fcdb}\underline{0.446} & \cellcolor[HTML]{fffbda}0.525 & 0.744 & 0.409 & 0.562 & 0.529 \\
 & Depth $L_1$ (cm) $\downarrow$ & \cellcolor[HTML]{99E4B0}\textbf{74.50} & \cellcolor[HTML]{99E4B0}\textbf{39.90} & \cellcolor[HTML]{99E4B0}\textbf{26.20} & \cellcolor[HTML]{99E4B0}\textbf{20.00} & \cellcolor[HTML]{99E4B0}\textbf{15.20} & \cellcolor[HTML]{99E4B0}\textbf{87.90} & \cellcolor[HTML]{99E4B0}\textbf{50.40} & \cellcolor[HTML]{99E4B0}\textbf{93.60} & \cellcolor[HTML]{99E4B0}\textbf{32.40} & \cellcolor[HTML]{99E4B0}\textbf{85.50} & \cellcolor[HTML]{99E4B0}\textbf{52.56} \\
 & PSNR-$\mu$ $\uparrow$ & \cellcolor[HTML]{fffbda}22.16 & \cellcolor[HTML]{fffbda}24.97 & \cellcolor[HTML]{fffbda}24.02 & 16.89 & \cellcolor[HTML]{fffbda}16.48 & \cellcolor[HTML]{fffbda}21.16 & \cellcolor[HTML]{fffbda}21.89 & 14.64 & \cellcolor[HTML]{fffbda}24.15 & \cellcolor[HTML]{fffbda}20.79 & \cellcolor[HTML]{fffbda}20.71 \\
\cdashline{1-13}
\rule{0pt}{2.2ex} \multirow{5}{*}{MonoGS (HDR)} & PSNR $\uparrow$ & \cellcolor[HTML]{fffbda}15.89 & \cellcolor[HTML]{d8fcdb}\underline{19.50} & \cellcolor[HTML]{fffbda}19.57 & \cellcolor[HTML]{99E4B0}\textbf{20.24} & \cellcolor[HTML]{d8fcdb}\underline{18.10} & \cellcolor[HTML]{fffbda}14.65 & \cellcolor[HTML]{d8fcdb}\underline{18.54} & 17.79 & \cellcolor[HTML]{fffbda}19.98 & \cellcolor[HTML]{d8fcdb}\underline{18.75} & \cellcolor[HTML]{fffbda}18.30 \\
 & SSIM $\uparrow$ & 0.454 & 0.598 & \cellcolor[HTML]{fffbda}0.565 & \cellcolor[HTML]{fffbda}0.474 & 0.391 & 0.417 & \cellcolor[HTML]{fffbda}0.500 & 0.264 & 0.645 & \cellcolor[HTML]{fffbda}0.494 & 0.480 \\
 & LPIPS $\downarrow$ & \cellcolor[HTML]{fffbda}0.505 & \cellcolor[HTML]{d8fcdb}\underline{0.394} & \cellcolor[HTML]{fffbda}0.528 & \cellcolor[HTML]{99E4B0}\textbf{0.467} & \cellcolor[HTML]{99E4B0}\textbf{0.496} & \cellcolor[HTML]{fffbda}0.532 & \cellcolor[HTML]{fffbda}0.525 & \cellcolor[HTML]{fffbda}0.654 & \cellcolor[HTML]{d8fcdb}\underline{0.356} & \cellcolor[HTML]{d8fcdb}\underline{0.482} & \cellcolor[HTML]{d8fcdb}\underline{0.494} \\
 & Depth $L_1$ (cm) $\downarrow$ & 194.04 & 135.51 & 90.67 & 64.11 & 65.14 & 252.26 & 139.82 & 214.16 & 109.33 & \cellcolor[HTML]{fffbda}170.58 & 143.56 \\
 & PSNR-$\mu$ $\uparrow$ & \cellcolor[HTML]{99E4B0}\textbf{23.89} & \cellcolor[HTML]{99E4B0}\textbf{26.11} & \cellcolor[HTML]{d8fcdb}\underline{25.00} & \cellcolor[HTML]{99E4B0}\textbf{22.79} & \cellcolor[HTML]{99E4B0}\textbf{20.52} & \cellcolor[HTML]{d8fcdb}\underline{22.58} & \cellcolor[HTML]{99E4B0}\textbf{22.95} & \cellcolor[HTML]{d8fcdb}\underline{19.07} & \cellcolor[HTML]{d8fcdb}\underline{24.19} & \cellcolor[HTML]{99E4B0}\textbf{24.12} & \cellcolor[HTML]{99E4B0}\textbf{23.12} \\
\cdashline{1-13}
\rule{0pt}{2.2ex} \multirow{5}{*}{Ours (HDR)} & PSNR $\uparrow$ & \cellcolor[HTML]{d8fcdb}\underline{16.89} & \cellcolor[HTML]{fffbda}19.02 & \cellcolor[HTML]{d8fcdb}\underline{20.96} & \cellcolor[HTML]{d8fcdb}\underline{19.80} & \cellcolor[HTML]{fffbda}18.02 & \cellcolor[HTML]{d8fcdb}\underline{18.09} & \cellcolor[HTML]{fffbda}18.24 & \cellcolor[HTML]{d8fcdb}\underline{18.59} & \cellcolor[HTML]{99E4B0}\textbf{22.10} & 17.50 & \cellcolor[HTML]{d8fcdb}\underline{18.92} \\
 & SSIM $\uparrow$ & \cellcolor[HTML]{d8fcdb}\underline{0.502} & \cellcolor[HTML]{fffbda}0.600 & \cellcolor[HTML]{d8fcdb}\underline{0.598} & \cellcolor[HTML]{fffbda}0.474 & \cellcolor[HTML]{fffbda}0.400 & \cellcolor[HTML]{d8fcdb}\underline{0.569} & 0.498 & \cellcolor[HTML]{fffbda}0.286 & \cellcolor[HTML]{d8fcdb}\underline{0.753} & 0.469 & \cellcolor[HTML]{fffbda}0.515 \\
 & LPIPS $\downarrow$ & \cellcolor[HTML]{99E4B0}\textbf{0.447} & \cellcolor[HTML]{99E4B0}\textbf{0.393} & \cellcolor[HTML]{99E4B0}\textbf{0.440} & \cellcolor[HTML]{d8fcdb}\underline{0.473} & \cellcolor[HTML]{d8fcdb}\underline{0.497} & \cellcolor[HTML]{99E4B0}\textbf{0.383} & \cellcolor[HTML]{d8fcdb}\underline{0.498} & \cellcolor[HTML]{d8fcdb}\underline{0.629} & \cellcolor[HTML]{99E4B0}\textbf{0.213} & \cellcolor[HTML]{fffbda}0.529 & \cellcolor[HTML]{99E4B0}\textbf{0.450} \\
 & Depth $L_1$ (cm) $\downarrow$ & \cellcolor[HTML]{fffbda}177.91 & \cellcolor[HTML]{fffbda}115.41 & \cellcolor[HTML]{fffbda}64.75 & \cellcolor[HTML]{d8fcdb}\underline{50.34} & \cellcolor[HTML]{fffbda}45.18 & \cellcolor[HTML]{fffbda}194.70 & \cellcolor[HTML]{fffbda}112.65 & \cellcolor[HTML]{d8fcdb}\underline{169.64} & \cellcolor[HTML]{fffbda}68.05 & 182.37 & \cellcolor[HTML]{fffbda}118.10 \\
 & PSNR-$\mu$ $\uparrow$ & \cellcolor[HTML]{d8fcdb}\underline{23.54} & \cellcolor[HTML]{d8fcdb}\underline{25.37} & \cellcolor[HTML]{99E4B0}\textbf{25.11} & \cellcolor[HTML]{d8fcdb}\underline{22.16} & \cellcolor[HTML]{d8fcdb}\underline{19.32} & \cellcolor[HTML]{99E4B0}\textbf{22.69} & \cellcolor[HTML]{d8fcdb}\underline{22.67} & \cellcolor[HTML]{99E4B0}\textbf{19.08} & \cellcolor[HTML]{99E4B0}\textbf{26.71} & \cellcolor[HTML]{d8fcdb}\underline{22.97} & \cellcolor[HTML]{d8fcdb}\underline{22.96} \\
\bottomrule
\end{tabular}%
}
\end{table*}

\begin{table*}[t]
\centering
\caption{Ablation study of different configuration components on LDR and HDR outputs per scene across 3 runs. The asterisk * marks that the baseline version of LDR includes a $L_{\text{SSIM}}$ term, while the vanilla HDR version does not include it.}
\label{tab:ablation_per_scene}
\renewcommand{\arraystretch}{1.2}
\resizebox{\linewidth}{!}{%
\begin{tabular}{cccccccccccccccc}
\hline
\multirow{2}{*}{\textbf{Image}} & \multicolumn{4}{c}{\textbf{Config}} & \multicolumn{11}{c}{\textbf{ATE RMSE} (cm)$\downarrow$} \\ \cline{2-16}
 & + Weight & + $L_{\text{SSIM}}$ & + L1-Reinhard & + log reparam. & bottles & boxes & cabin & candles & christmas & coat rack & coffee & kitchen & nerdy robot & small city & \textbf{Mean} \\ \hline
\multirow{8}{*}{LDR} & \makebox[1.2em][c]{\xmark} & \makebox[1.2em][c]{\textbf{\cmark}} & \makebox[1.2em][c]{\xmark} & \makebox[1.2em][c]{\xmark} & 52.61 & 48.30 & 23.56 & 74.22 & 38.76 & \cellcolor[HTML]{fffbda}60.99 & 74.85 & \cellcolor[HTML]{fffbda}37.06 & 68.39 & 89.61 & 56.84 \\
 & \makebox[1.2em][c]{\xmark} & \makebox[1.2em][c]{\cmark} & \makebox[1.2em][c]{\xmark} & \makebox[1.2em][c]{\textbf{\cmark}} & 39.25 & 48.11 & 10.09 & 36.31 & 39.18 & 73.24 & 33.18 & 38.65 & 50.88 & 43.04 & 41.19 \\
 & \makebox[1.2em][c]{\xmark} & \makebox[1.2em][c]{\cmark} & \makebox[1.2em][c]{\textbf{\cmark}} & \makebox[1.2em][c]{\xmark} & 28.54 & \cellcolor[HTML]{FCDFD8}\underline{24.29} & \cellcolor[HTML]{fffbda}7.11 & \cellcolor[HTML]{FCDFD8}\underline{18.06} & 43.12 & \cellcolor[HTML]{FCDFD8}\underline{60.98} & \cellcolor[HTML]{fffbda}26.50 & 43.94 & 14.63 & 33.95 & 30.11 \\
 & \makebox[1.2em][c]{\xmark} & \makebox[1.2em][c]{\cmark} & \makebox[1.2em][c]{\textbf{\cmark}} & \makebox[1.2em][c]{\textbf{\cmark}} & \cellcolor[HTML]{FCDFD8}\underline{27.84} & 31.48 & 7.17 & \cellcolor[HTML]{fffbda}20.16 & \cellcolor[HTML]{FCDFD8}\underline{33.98} & \cellcolor[HTML]{FF9999}\textbf{59.19} & \cellcolor[HTML]{FCDFD8}\underline{26.18} & 45.65 & 16.29 & 38.02 & 30.60 \\ \cdashline{2-16}
 & \makebox[1.2em][c]{\xmark} & \makebox[1.2em][c]{\xmark} & \makebox[1.2em][c]{\textbf{\cmark}} & \makebox[1.2em][c]{\textbf{\cmark}} & 30.05 & \cellcolor[HTML]{fffbda}26.06 & 7.40 & \cellcolor[HTML]{FF9999}\textbf{14.39} & 40.96 & 66.81 & \cellcolor[HTML]{FF9999}\textbf{23.56} & 38.60 & \cellcolor[HTML]{fffbda}14.49 & \cellcolor[HTML]{FF9999}\textbf{25.24} & \cellcolor[HTML]{fffbda}28.76 \\
 & \makebox[1.2em][c]{\textbf{\cmark}} & \makebox[1.2em][c]{\xmark} & \makebox[1.2em][c]{\textbf{\cmark}} & \makebox[1.2em][c]{\textbf{\cmark}} & \cellcolor[HTML]{fffbda}27.89 & \cellcolor[HTML]{FF9999}\textbf{23.87} & \cellcolor[HTML]{FCDFD8}\underline{6.88} & 21.43 & \cellcolor[HTML]{fffbda}36.80 & 64.99 & 30.38 & \cellcolor[HTML]{FCDFD8}\underline{31.67} & \cellcolor[HTML]{FCDFD8}\underline{9.59} & \cellcolor[HTML]{fffbda}33.22 & \cellcolor[HTML]{FCDFD8}\underline{28.67} \\
 & \makebox[1.2em][c]{\xmark} & \makebox[1.2em][c]{\textbf{\cmark}} & \makebox[1.2em][c]{\textbf{\cmark}} & \makebox[1.2em][c]{\textbf{\cmark}} & \cellcolor[HTML]{FCDFD8}\underline{27.84} & 31.48 & 7.17 & \cellcolor[HTML]{fffbda}20.16 & \cellcolor[HTML]{FCDFD8}\underline{33.98} & \cellcolor[HTML]{FF9999}\textbf{59.19} & \cellcolor[HTML]{FCDFD8}\underline{26.18} & 45.65 & 16.29 & 38.02 & 30.60 \\
 & \makebox[1.2em][c]{\textbf{\cmark}} & \makebox[1.2em][c]{\textbf{\cmark}} & \makebox[1.2em][c]{\textbf{\cmark}} & \makebox[1.2em][c]{\textbf{\cmark}} & \cellcolor[HTML]{FF9999}\textbf{27.00} & 31.53 & \cellcolor[HTML]{FF9999}\textbf{6.40} & 23.59 & \cellcolor[HTML]{FF9999}\textbf{29.89} & 61.24 & 26.83 & \cellcolor[HTML]{FF9999}\textbf{16.07} & \cellcolor[HTML]{FF9999}\textbf{7.36} & \cellcolor[HTML]{FCDFD8}\underline{29.82} & \cellcolor[HTML]{FF9999}\textbf{25.97} \\ \hline
 \multirow{8}{*}{HDR} & \makebox[1.2em][c]{\xmark} & \makebox[1.2em][c]{\xmark} & \makebox[1.2em][c]{\xmark} & \makebox[1.2em][c]{\xmark} & 61.20 & 34.06 & 42.54 & 57.98 & 36.75 & \cellcolor[HTML]{d8fcdb}\underline{63.71} & 51.37 & 40.58 & 59.85 & 46.25 & 49.43 \\
 & \makebox[1.2em][c]{\xmark} & \makebox[1.2em][c]{\xmark} & \makebox[1.2em][c]{\xmark} & \makebox[1.2em][c]{\textbf{\cmark}} & 40.17 & 24.31 & 38.05 & 44.67 & 37.80 & 68.80 & 45.30 & \cellcolor[HTML]{fffbda}24.94 & 22.69 & 38.00 & 38.47 \\
 & \makebox[1.2em][c]{\xmark} & \makebox[1.2em][c]{\xmark} & \makebox[1.2em][c]{\textbf{\cmark}} & \makebox[1.2em][c]{\xmark} & 28.66 & \cellcolor[HTML]{fffbda}23.39 & 13.74 & 19.13 & 41.09 & 66.51 & 28.88 & 30.13 & 9.97 & \cellcolor[HTML]{d8fcdb}\underline{23.42} & 28.49 \\
 & \makebox[1.2em][c]{\xmark} & \makebox[1.2em][c]{\xmark} & \makebox[1.2em][c]{\textbf{\cmark}} & \makebox[1.2em][c]{\textbf{\cmark}} & \cellcolor[HTML]{d8fcdb}\underline{27.43} & 26.03 & \cellcolor[HTML]{fffbda}10.92 & \cellcolor[HTML]{fffbda}17.59 & \cellcolor[HTML]{d8fcdb}\underline{26.34} & \cellcolor[HTML]{99E4B0}\textbf{60.74} & 29.01 & 37.00 & 8.02 & \cellcolor[HTML]{fffbda}24.32 & \cellcolor[HTML]{d8fcdb}\underline{26.74} \\ \cdashline{2-16}
 & \makebox[1.2em][c]{\xmark} & \makebox[1.2em][c]{\xmark} & \makebox[1.2em][c]{\textbf{\cmark}} & \makebox[1.2em][c]{\textbf{\cmark}} & \cellcolor[HTML]{d8fcdb}\underline{27.43} & 26.03 & \cellcolor[HTML]{fffbda}10.92 & \cellcolor[HTML]{fffbda}17.59 & \cellcolor[HTML]{d8fcdb}\underline{26.34} & \cellcolor[HTML]{99E4B0}\textbf{60.74} & 29.01 & 37.00 & 8.02 & \cellcolor[HTML]{fffbda}24.32 & \cellcolor[HTML]{d8fcdb}\underline{26.74} \\
 & \makebox[1.2em][c]{\textbf{\cmark}} & \makebox[1.2em][c]{\xmark} & \makebox[1.2em][c]{\textbf{\cmark}} & \makebox[1.2em][c]{\textbf{\cmark}} & \cellcolor[HTML]{99E4B0}\textbf{24.46} & 42.74 & \cellcolor[HTML]{d8fcdb}\underline{10.68} & \cellcolor[HTML]{99E4B0}\textbf{10.20} & \cellcolor[HTML]{fffbda}31.98 & \cellcolor[HTML]{fffbda}65.48 & \cellcolor[HTML]{fffbda}24.58 & \cellcolor[HTML]{d8fcdb}\underline{21.45} & \cellcolor[HTML]{d8fcdb}\underline{6.85} & 38.31 & \cellcolor[HTML]{fffbda}27.67 \\
 & \makebox[1.2em][c]{\xmark} & \makebox[1.2em][c]{\textbf{\cmark}} & \makebox[1.2em][c]{\textbf{\cmark}} & \makebox[1.2em][c]{\textbf{\cmark}} & 32.04 & \cellcolor[HTML]{d8fcdb}\underline{19.05} & 11.05 & 22.34 & 41.34 & 69.40 & \cellcolor[HTML]{d8fcdb}\underline{24.39} & 49.23 & \cellcolor[HTML]{fffbda}7.11 & \cellcolor[HTML]{99E4B0}\textbf{22.88} & 29.88 \\
 & \makebox[1.2em][c]{\textbf{\cmark}} & \makebox[1.2em][c]{\textbf{\cmark}} & \makebox[1.2em][c]{\textbf{\cmark}} & \makebox[1.2em][c]{\textbf{\cmark}} & \cellcolor[HTML]{fffbda}27.91 & \cellcolor[HTML]{99E4B0}\textbf{15.19} & \cellcolor[HTML]{99E4B0}\textbf{9.97} & \cellcolor[HTML]{d8fcdb}\underline{15.58} & \cellcolor[HTML]{99E4B0}\textbf{25.70} & 71.18 & \cellcolor[HTML]{99E4B0}\textbf{23.15} & \cellcolor[HTML]{99E4B0}\textbf{17.14} & \cellcolor[HTML]{99E4B0}\textbf{6.69} & 29.02 & \cellcolor[HTML]{99E4B0}\textbf{24.15} \\ \hline
\end{tabular}%
}
\end{table*}

\end{document}